\documentclass[10pt,twocolumn,letterpaper]{article}

\usepackage{cvpr}      

\usepackage{graphicx}
\usepackage{amsmath}
\usepackage{amssymb}
\usepackage{booktabs}
\usepackage[frozencache,cachedir=.]{minted}
\usepackage[accsupp]{axessibility}
\makeatletter
\@namedef{ver@everyshi.sty}{}
\makeatother
\usepackage{tikz}
\usepackage[skins]{tcolorbox}
\usetikzlibrary{positioning}
\usepackage{pgfplots}
\usetikzlibrary{shapes.geometric}
\usetikzlibrary{arrows.meta,arrows}
\usepackage{multirow}
\usepackage[pagebackref,breaklinks,colorlinks]{hyperref}

\usepackage[capitalize]{cleveref}

\def\cvprPaperID{56} 
\def\confName{EarthVision}
\def\confYear{2022}

\newcommand{\orionsar}{{H}ephaestus}

\begin{document}

\title{\orionsar{}: A large scale multitask dataset towards InSAR understanding}

\author{Nikolaos Ioannis Bountos$^{1,2}$ \and Ioannis Papoutsis$^1$ \and Dimitrios Michail$^2$ \and Andreas Karavias$^3$ \and Panagiotis Elias$^1$ \and Isaak Parcharidis$^3$ \and 
$^1$Institute for Astronomy, Astrophysics, Space Applications, and Remote Sensing,\\ National Observatory of Athens\\
$^2$Department of Informatics and Telematics, Harokopio University of Athens\\
$^3$Department of Geography, Harokopio University of Athens\\
{\tt\small \{bountos,ipapoutsis,pelias\}@noa.gr, \{michail,karavias,parchar\}@hua.gr}
}

\maketitle

\begin{abstract}
Synthetic Aperture Radar (SAR) data and Interferometric SAR (InSAR) products in particular, are one of the largest sources of Earth Observation data. InSAR provides unique information on diverse geophysical processes and geology, and on the geotechnical properties of man-made structures. However, there are only a limited number of applications that exploit the abundance of InSAR data and deep learning methods to extract such knowledge. The main barrier has been the lack of a large curated and annotated InSAR dataset, which would be costly to create and would require an interdisciplinary team of experts experienced on InSAR data interpretation. In this work, we put the effort to create and make available the first of its kind, manually annotated dataset that consists of 19,919 individual Sentinel-1 interferograms acquired over 44 different volcanoes globally, which are split into 216,106 InSAR patches. The annotated dataset is designed to address different computer vision problems, including volcano state classification, semantic segmentation of ground deformation, detection and classification of atmospheric signals in InSAR imagery, interferogram captioning, text to InSAR generation, and InSAR image quality assessment.

\end{abstract}

\section{Introduction}
\label{sec:intro}

The Copernicus program is believed to be a game changer for earth system sciences.
Sentinel-1 Interferometric Synthetic Aperture Radar (InSAR) data, typically generated by subtracting the phase of two SAR scenes, the primary and the secondary, acquired at two different points in time, has evolved from use-case specific applications to routine global monitoring over yearly time spans~\cite{biggs2020satellite}. 

Among other applications, differential InSAR (DInSAR - used interchangeably with terms InSAR and interferogram in this paper) encapsulates rich and diverse information that has been widely used for modeling earthquakes and studying tectonics~\cite{boncori2015february}, understanding magmatism and mitigating volcanic hazards~\cite{2012GL054137}, monitoring surface motion due to anthropogenic or physical driving mechanisms at local~\cite{el2021monitoring} or wide area scales~\cite{rs12193207}, and glacier change detection~\cite{strozzi2020monitoring}. 

The availability of a huge archive of Sentinel-1 data, to which new SAR images are systematically added at a rate of 3.5 TB per day, makes InSAR an excellent fit for computer vision tasks, addressing the thematic applications above. However, despite the extensive use of deep learning (DL) in remote sensing~\cite{zhu2017deep}, research has focused more on multi-spectral modalities and lately on SAR backscatter~\cite{zhu2021deep}. There are far fewer applications that use DL to exploit the geodetic nature of InSAR, i.e. the phase component of SAR imagery. Previous works rely on Convolutional Neural Networks (CNN) and supervised learning with few training samples, for single, wrapped or unwrapped, interferogram classification. These approaches rely on heavy data augmentation and/or synthetic data generation~\cite{anantrasirichai2018application, gaddesasimultaneous, 2019GL084993, bountos2021self, bountos2022learning, valade2019towards, brengman2021identification} to cope with class imbalance and the scarcity of training data. Similar DL approaches have also been tested on time-series of InSAR data~\cite{2019GL084993, sun2020automatic}. To circumvent the lack of training data, Bountos \etal ~\cite{bountos2021self} proposed a self-supervised learning approach to create quality InSAR feature extractors, making use of a much larger archive of  real, unlabeled, InSAR data.

The InSAR community is currently lacking a large, curated and annotated InSAR dataset, which would spur and scale the development of new applications for geophysical research and geohazards mitigation. 
However, creating a carefully annotated InSAR dataset is a non-trivial task, that is hard to automate. SAR interferograms are unique in their nature and consist of a superposition of several signals, ranging from ground deformations due to different geophysical and/or man-made mechanisms, atmospheric disturbances, digital elevation model corrections, orbital errors, temporal, land cover and other signal decorrelation factors appearing as noise, etc. Therefore InSAR data annotation requires InSAR experts that are able to identify the subtle differences in each InSAR, interpret and annotate interferograms considering the geological, tectonic, meteorological and InSAR imaging context. 

In this work, 
we design and generate \orionsar{}, a fine-grained, manually annotated InSAR dataset focused around global volcano monitoring. To the best of our knowledge this is the first annotated InSAR dataset of such scale, designed to address multiple computer vision tasks. With a team of InSAR experts consisting of Earth Observation scientists, geologists, geophysicists, and computer scientists, we annotate 19,919 individual interferograms and provide a diverse set of labels about 
atmospheric contributions, ground deformation fringes, various details for the event under study \eg its intensity and the state of the volcano, the quality of the interferogram as well as a textual description of the interferogram itself. \orionsar{} is therefore labeled with a strategy to tackle different machine learning problems, including multi-label multi-class InSAR classification, semantic segmentation, image quality assessment, image captioning, and synthetic InSAR data generation.        
Along with the annotated InSAR set, we provide a large, unlabeled dataset with 110,573 interferograms to research methods for the creation of general InSAR foundation models applicable to multiple downstream tasks not covered in this work. The dataset and the code used in this work is published and maintained on: 
\url{https://github.com/Orion-AI-Lab/Hephaestus}.

\section{Dataset Description} 
Our dataset revolves around the 44 most active volcanoes globally. For these volcanoes we collected wrapped DInSAR data, i.e. phase values are modulo 2$\pi$, from the Comet-LiCS portal~\cite{lazecky2020licsar,morishita2020licsbas,wright2016licsar,lawrence2013storing}, which corresponds to 38 different descending and ascending Sentinel-1 A\&B frames, covering the period 2014-2021. One frame may cover more than one volcano. 
All InSAR products are accompanied with the corresponding coherence maps. 

\subsection{Volcanoes under study}
\begin{figure}
\centering
    \includegraphics[width=0.5\textwidth]{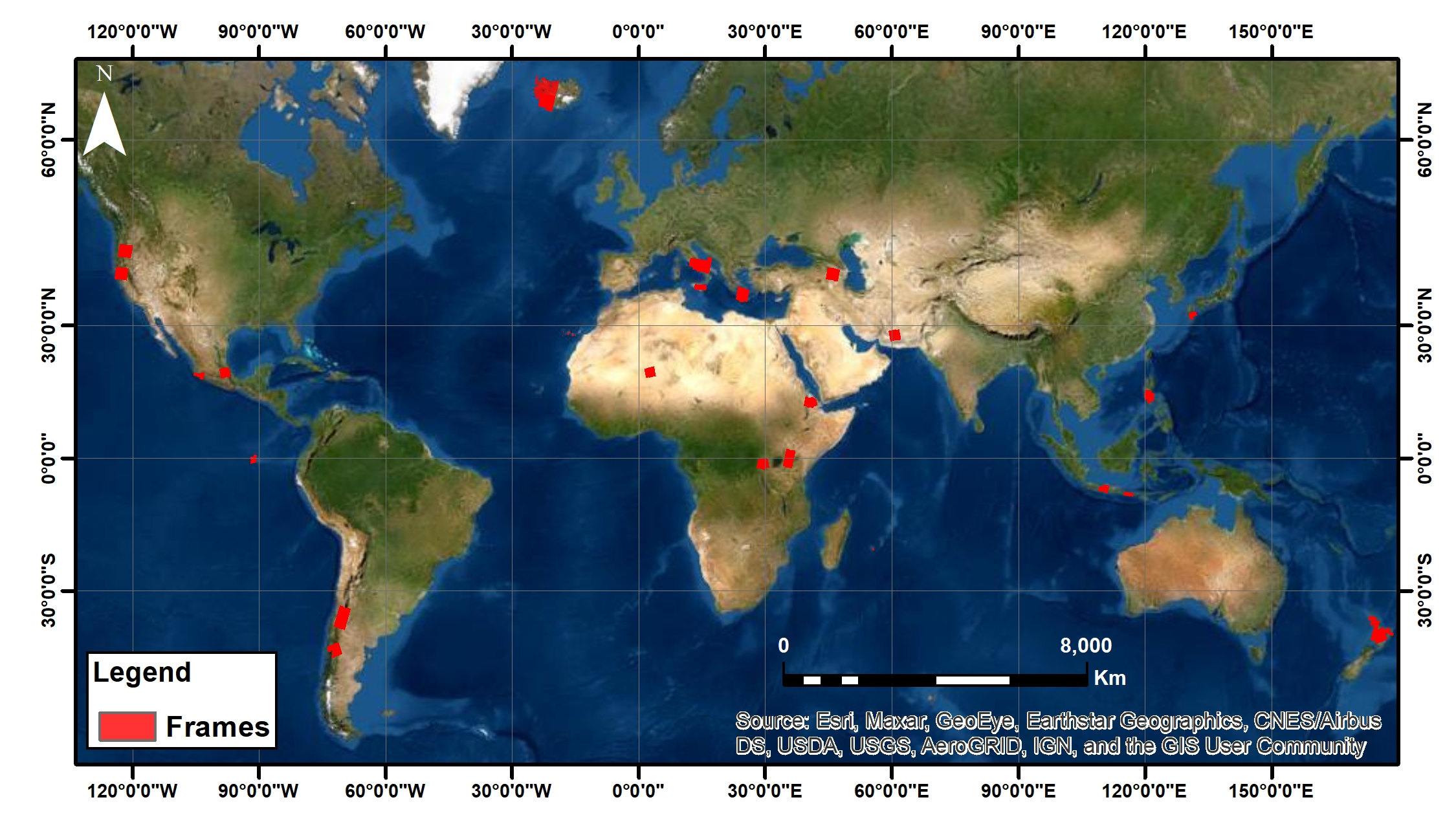}
    \includegraphics[width=0.5\textwidth]{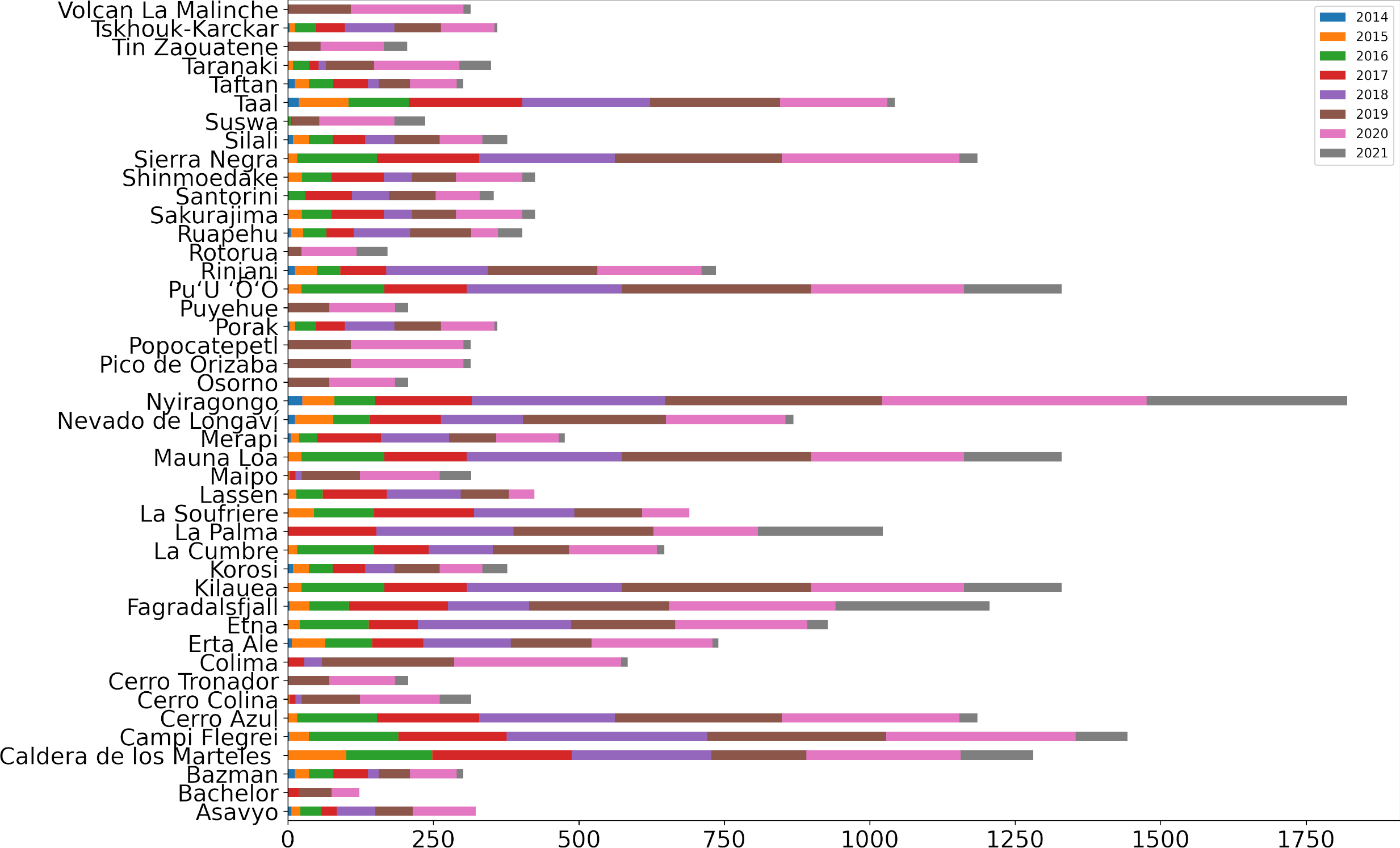}
    \caption{Spatio-temporal distribution of the InSAR frames in \orionsar{}. The names of the volcanoes appearing in the dataset are shown at the y-axis.}
\label{fig:volcanoes}
\end{figure}

\Cref{fig:volcanoes} shows the volcanoes included in \orionsar{} and the temporal distribution of InSAR frames used for each volcano. We collected all available Sentinel-1 InSAR from 2014 to 2021 summing up to 19,919 samples from which 1,833 contain ground deformation due to volcanic activity or earthquake. Almost all of the major volcanic events in that period were covered, including volcanoes with recent activity, such as Etna, Taal, La Palma and Fagradalsfjall. 

\subsection{Annotation strategy}
\label{sec:annotation}
The manual annotation of the dataset is challenging, since it involves the correct interpretation of interferometric fringes and their separation into atmospheric effects and actual deformation, the classification of observed deformation into volcanic activity or earthquake event, and the categorization of the volcanic deformation according to geodetic models. The assessment and interpretation of the impact of atmospheric disturbances is a particularly difficult task, since their InSAR signature may give rise to false alarms of volcanic deformation, due to the look-alike patterns that are mainly caused by vertical stratification effects on high altitude areas like mountains or volcanoes~\cite{parker2015systematic}. In order to avoid such false alarms, our experts exploited extensive literature sources about the historic activity of each volcano and their InSAR interpretation skills.

For the manual annotation of the dataset, twenty different label categories were selected to characterize every interferogram. An example of such annotation can be seen in \cref{mint:json}. The first four labels are metadata like the \textit{uniqueID}, which is a unique code to identify every image, the \textit{frame ID} to determine the satellite orbit and frame, and  the \textit{primary date} and \textit{secondary date} for each SAR image acquisition. The next family of labels corresponds to technical errors that may take place during the automatic InSAR generation by Comet-LiCS. First, is the the binary category \textit{corrupted}, that separates interferograms that are totally problematic. Second, is the category \textit{processing error} that refers to two different InSAR processing error types: \textit{type 1} refers to debursting error during the synchronization of the burst of one or more sub-swaths of the images, \textit{type 2} refers to Sentinel-1 sub-swath merging errors that appear as a visual discontinuity. \textit{Type 0} is used when there are no processing errors appearing on the interferogram. 

The next family of labels specify characteristics and patterns with a value of \textit{0} when these are not present in the InSAR and \textit{1} when they are. In the case of multiple subclasses, each number greater than \textit{0} will denote one subclass.
These labels are  a) \textit{glacier fringes} when observed fringes are caused by glacier melting, b) \textit{orbital fringes} when the wrapped interferograms present a phase ramp due to orbital errors, c) \textit{atmospheric fringes} for which we have four types of InSAR atmospheric effect classes; \textit{type 0} when there is no atmospheric impact, \textit{type 1} when there is vertical stratification that is correlated with topography and is caused by changes of the refractive index of the troposphere, \textit{type 2} when there is turbulent mixing and vapors caused by liquid and solid particles of the atmosphere\cite{hanssen1998atmospheric}, and \textit{type 3} when both effects (type 1,2) are interleaved, d) \textit{low coherence} when the images are characterized from interferometric signal decorrelation, e) \textit{no info} indicating low coherence throughout the entire interferogram, making it hard to extract information, and f) \textit{image artifacts} which point to an artificial colorbar legend on the image.

The final label family describes the observed deformation, associated with volcanic activity. The field \textit{label} refers to the existence of deformation and can be \textit{Non Deformation}, \textit{Deformation}, and \textit{Earthquake}. \textit{Activity type} provides a classification of the volcanic deformation according to magma sources in \textit{Mogi}~\cite{Mogi1958}, \textit{Dyke}\cite{okada1985surface,sigmundsson2010intrusion}, \textit{Sill}~\cite{fialko2001deformation}, \textit{Spheroid}~\cite{yang1988deformation} and \textit{Unidentified}. The \textit{intensity level} category discriminates the intensity of displacement caused by volcanic activity and can be \textit{None}, \textit{Low} (one observed fringe), \textit{Medium} (two or three fringes), and \textit{High} (more than three fringes). The \textit{phase} category classifies periods of volcanic activity based on the pattern of the fringes to \textit{Rest}, \textit{Unrest} (uplift is observed) and \textit{Rebound} (subsidence is observed). An important label is \textit{confidence}, which determines the level of uncertainty during the annotation for the deformation categories (values in the range from 0 to 1). \textit{Is crowd} has a value of \textit{0} when there is maximum one local fringe pattern in the interferogram, and \textit{1} when two or more such patterns exist in the same InSAR image. Furthermore, we provide a manually drawn \textit{segmentation mask}, which specifies the displacement area.
Finally a description of the interferogram is provided in the \textit{caption} field.

\begin{listing}
\begin{minted}[fontsize=\footnotesize]{json}
{
  "uniqueID": 1957,
  "frameID": "124D_05291_081406",
  "primary_date": "20200110",
  "secondary_date": "20200122",
  "corrupted": 0,
  "processing_error": 0,
  "glacier_fringes": 0,
  "orbital_fringes": 0,
  "atmospheric_fringes": 1,
  "low_coherence": 0,
  "no_info": 0,
  "image_artifacts": 0,
  "label": [ "Deformation" ],
  "activity_type": [ "Mogi" ],
  "intensity_level": "Medium",
  "phase": "Unrest",
  "confidence": 0.6,
  "segmentation_mask": [
    [
      616.1845703125,
      121.2841796875,
      ...,
      618.5218671292696,
      116.14156159975028
    ]
  ],
  "is_crowd": 0,
  "caption": "Vertical stratification caused
  by atmospheric delays can be detected in high
  altitude areas. Medium deformation activity
  can be detected."
}
\end{minted}
\caption{Annotation file example for one \orionsar{} sample.}\label{mint:json}
\end{listing}

\subsection{Data Distribution}
\orionsar{} is highly diverse as we have
included wrapped InSAR 1) from both descending
and ascending viewing geometries, 2) produced with different range-azimuth looks combinations, 3) with pure phase-only information and with the phase overlaid with SAR amplitude, and 4) produced with different color scale palettes~\cite{lazecky2020licsar}.

\begin{figure*}[]
\centering
    \begin{subfigure}{0.45\textwidth}
    \centering
    \includegraphics[width=\textwidth]{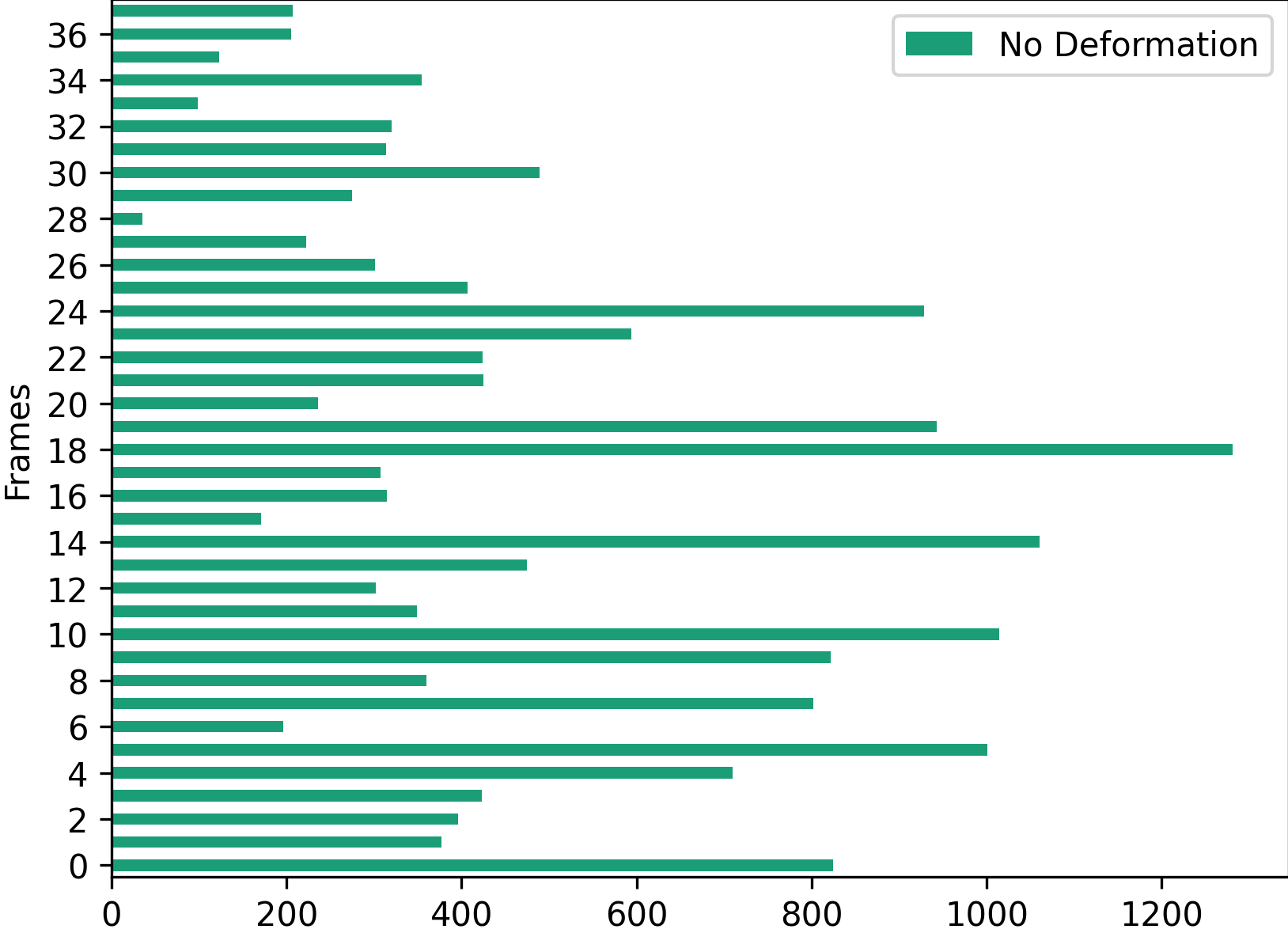}
      \caption{}
      \label{fig:distra}
    \end{subfigure}
   \begin{subfigure}{0.45\textwidth}
   \centering
   \includegraphics[width=\textwidth]{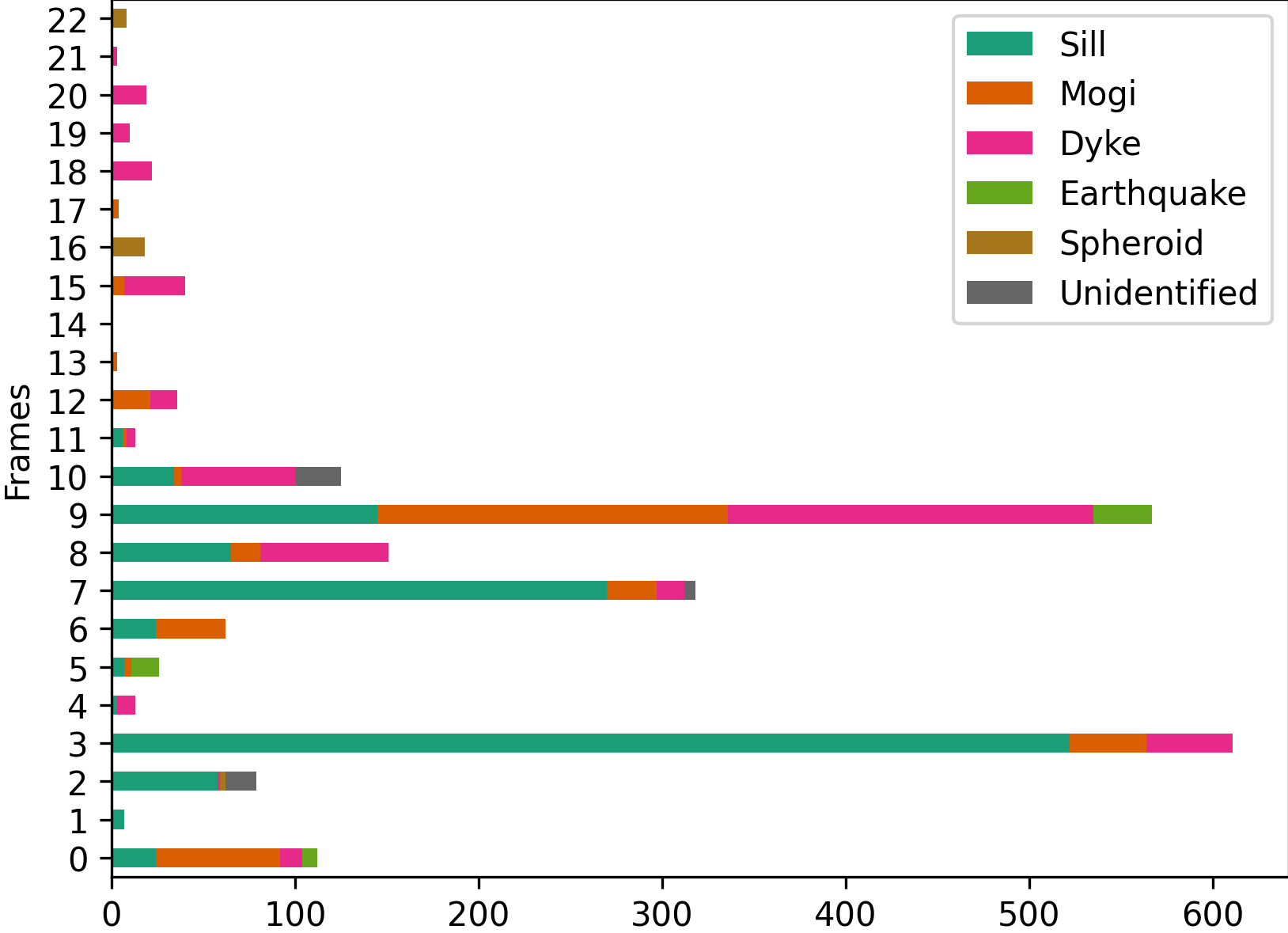}
   \caption{}
      \label{fig:distrb}
   \end{subfigure}
\caption{Distribution of classes among frames. a) contains the frames with no deformation while b) the frames with deformation. The frameIDs are omitted for this plot.}
\label{fig:class_distribution}
\end{figure*}

\begin{figure}
\centering
    \includegraphics[width=0.45\textwidth]{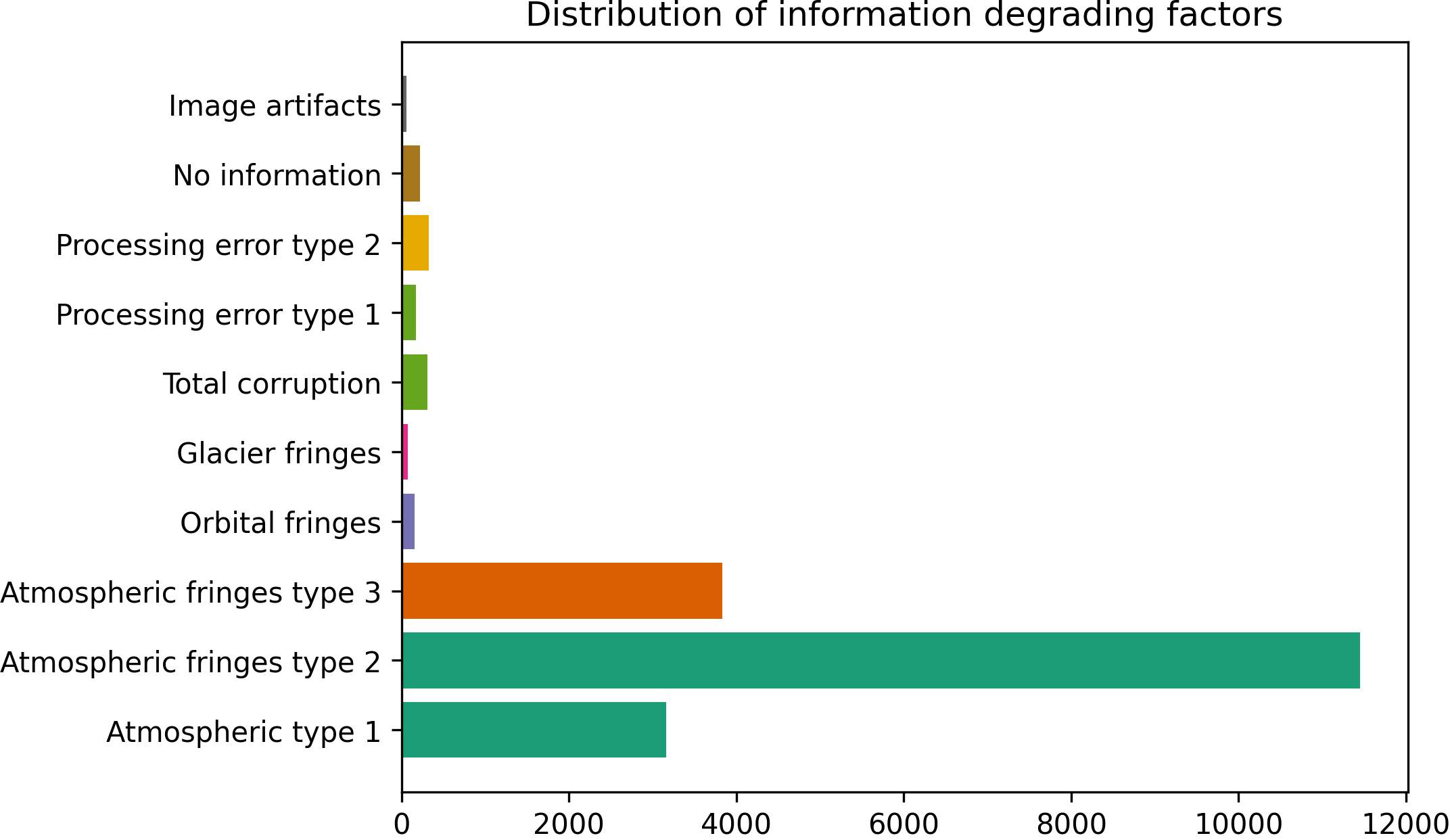}
    \caption{Distribution of labels related to InSAR information degradation.}\label{fig:disturbances}
\end{figure}

In \cref{fig:volcanoes} we show the spatiotemporal distribution of the records based on the date of the primary SAR image. 
In \cref{fig:class_distribution}, we show the distribution of the samples based on the 38 frame ids; \Cref{fig:distra} contains the samples without deformation and \cref{fig:distrb} the samples with deformation with different volcanic mechanisms. 
There is a major class imbalance between ground deformation and no deformation classes. This is anticipated since volcanic activity and earthquakes are not the norm in nature. In general, natural disaster related problems such as fire risk prediction~\cite{prapas2021deep} and volcanic activity detection~\cite{bountos2021self} are closely tied with long tailed distributions. Taking a deeper look at the deformation samples, \Cref{tab:def_distr,tab:phase} present a breakdown in regards to deformation type, intensity of the event and phase of the volcano. Again, some of the classes are imbalanced. Most of the recorded volcanic episodes are in the unrest phase and classified as Mogi, Dyke or Sill. 
Expanding the dataset to contain a greater amount of frames will not solve the class imbalance problem since the number of unrest events is finite in the period 2014-2021.
Finally, the distribution of samples containing atmospheric signals, background noise and other processing errors that degrade InSAR quality are shown in~\Cref{fig:disturbances}.

\begin{table}
    \centering

    \begin{tabular}{cccc}

    \toprule
        Deformation Type & & Intensity & \\
        \midrule
         & Low & Medium & High\\
         \midrule
        Mogi & 196 & 111 & 72 \\
        \hline
        Dyke & 64 & 79 & 322 \\
        \hline
        Sill&  492 & 332 & 320\\ \hline
        Spheroid&19& 6 & 3\\ \hline
        Earthquake&0 & 0 &40\\ \hline
        Unidentified & 14 & 4 &30\\ 
        \bottomrule
    \end{tabular}
    \caption{Deformation type vis-à-vis intensity breakdown.}
    \label{tab:def_distr}

\end{table}

\begin{table}[]
\centering
\begin{tabular}{lll}
\toprule
Deformation Type & \multicolumn{2}{l}{Phase} \\
\midrule
                 & Unrest      & Rebound     \\ \midrule
Mogi             &  283           &       96      \\ \hline
Dyke             &  462           &     3        \\ \hline
Sill             &  1102           &    42         \\ \hline
Spheroid         &   28          &      0       \\ \hline
Earthquake       &  40           &  15           \\ \hline
Unidentified     &  41           &  7         \\ 
\bottomrule
\end{tabular}
\caption{Deformation type vis-à-vis volcano phase breakdown.}
\label{tab:phase}
\end{table}

\subsection{Time-series nature of the dataset}
\label{sec:spatiotemporal}
\begin{figure*}
    \centering
    \begin{subfigure}{0.45\textwidth}
    \centering
    \includegraphics[width=\textwidth]{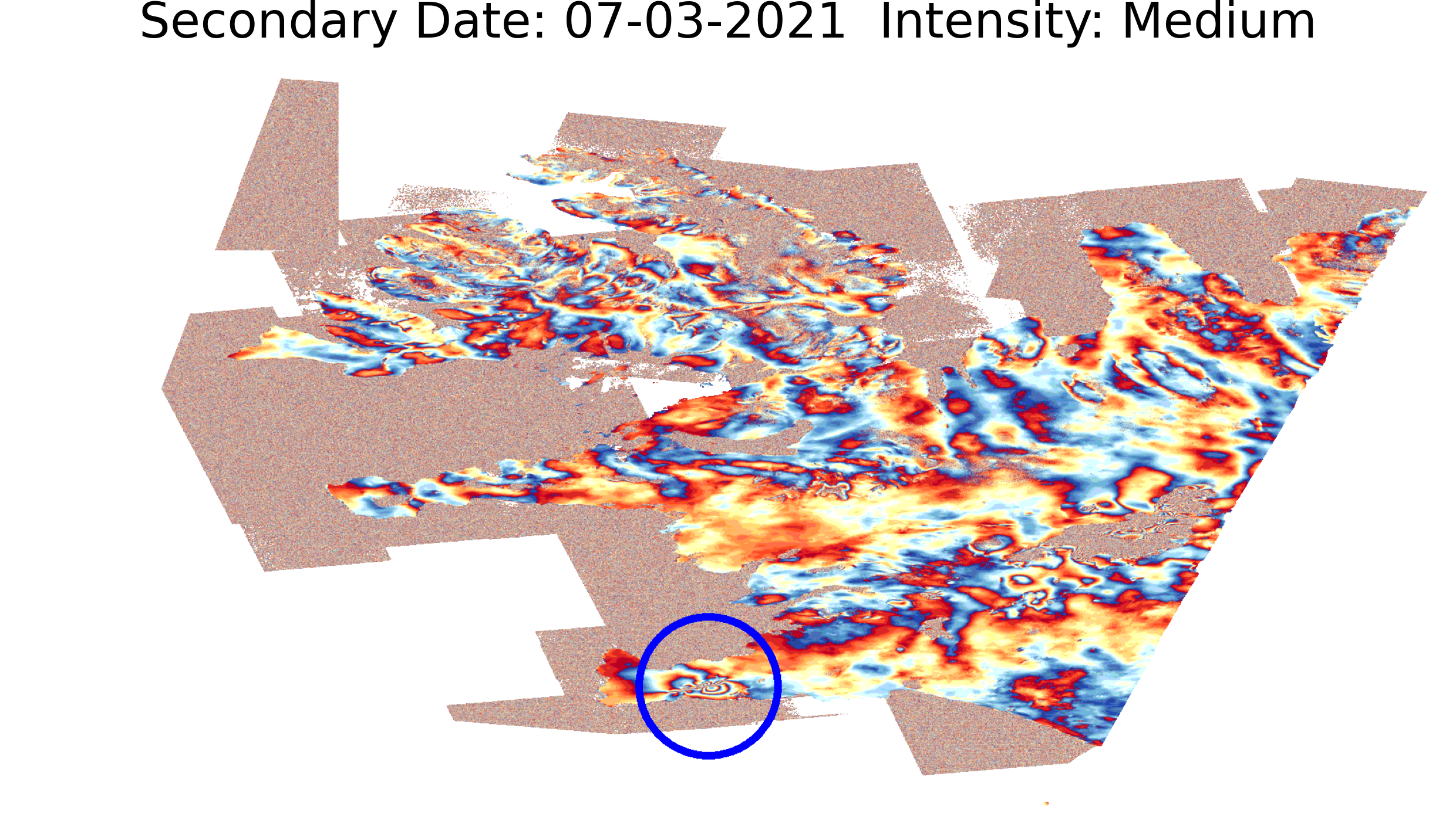}
    \end{subfigure}
    \begin{subfigure}{0.45\textwidth}
    \centering
    \includegraphics[width=\textwidth]{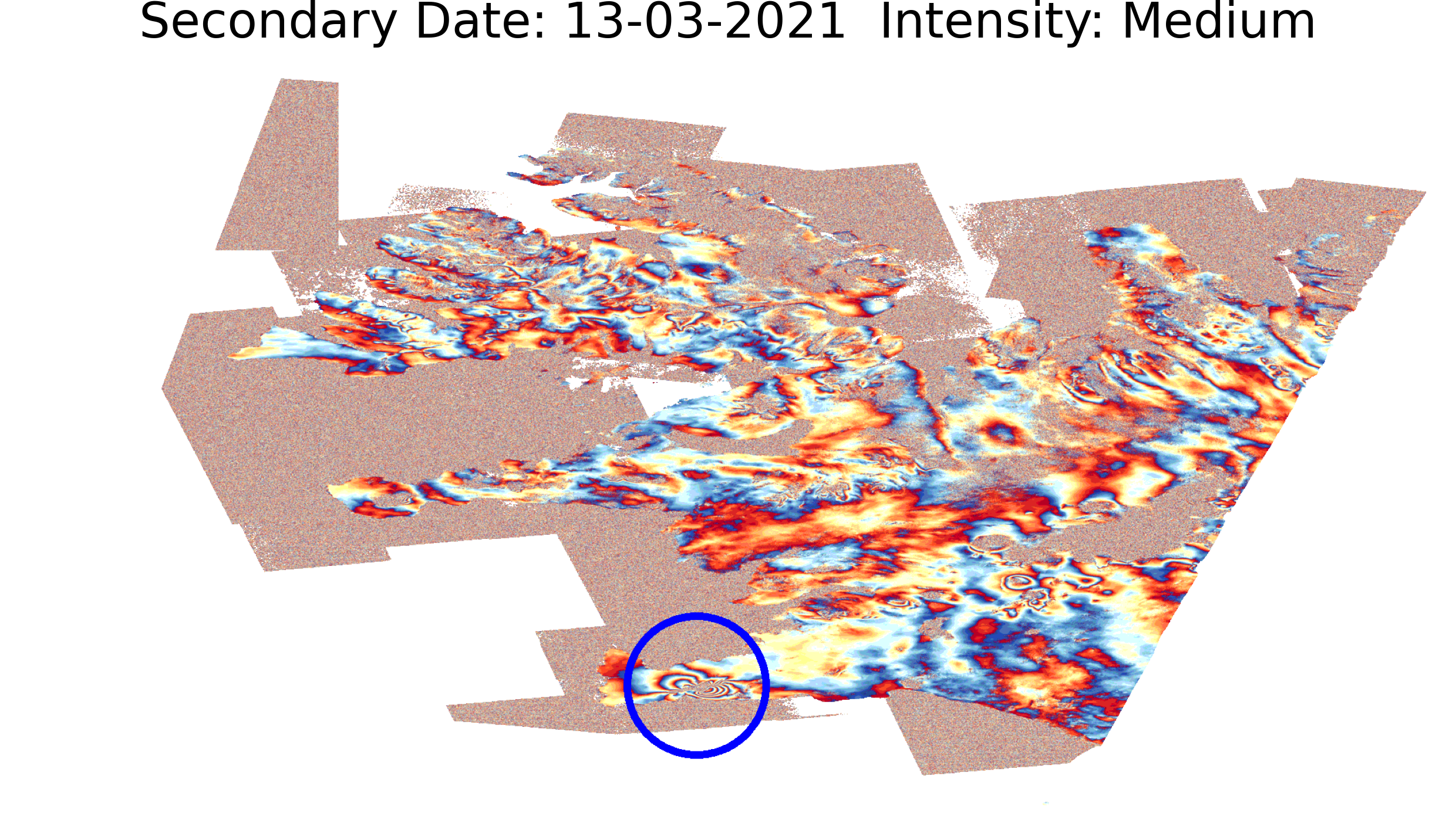}
    \end{subfigure}
    \hfill
    \begin{subfigure}{0.45\textwidth}
    \centering
    \includegraphics[width=\textwidth]{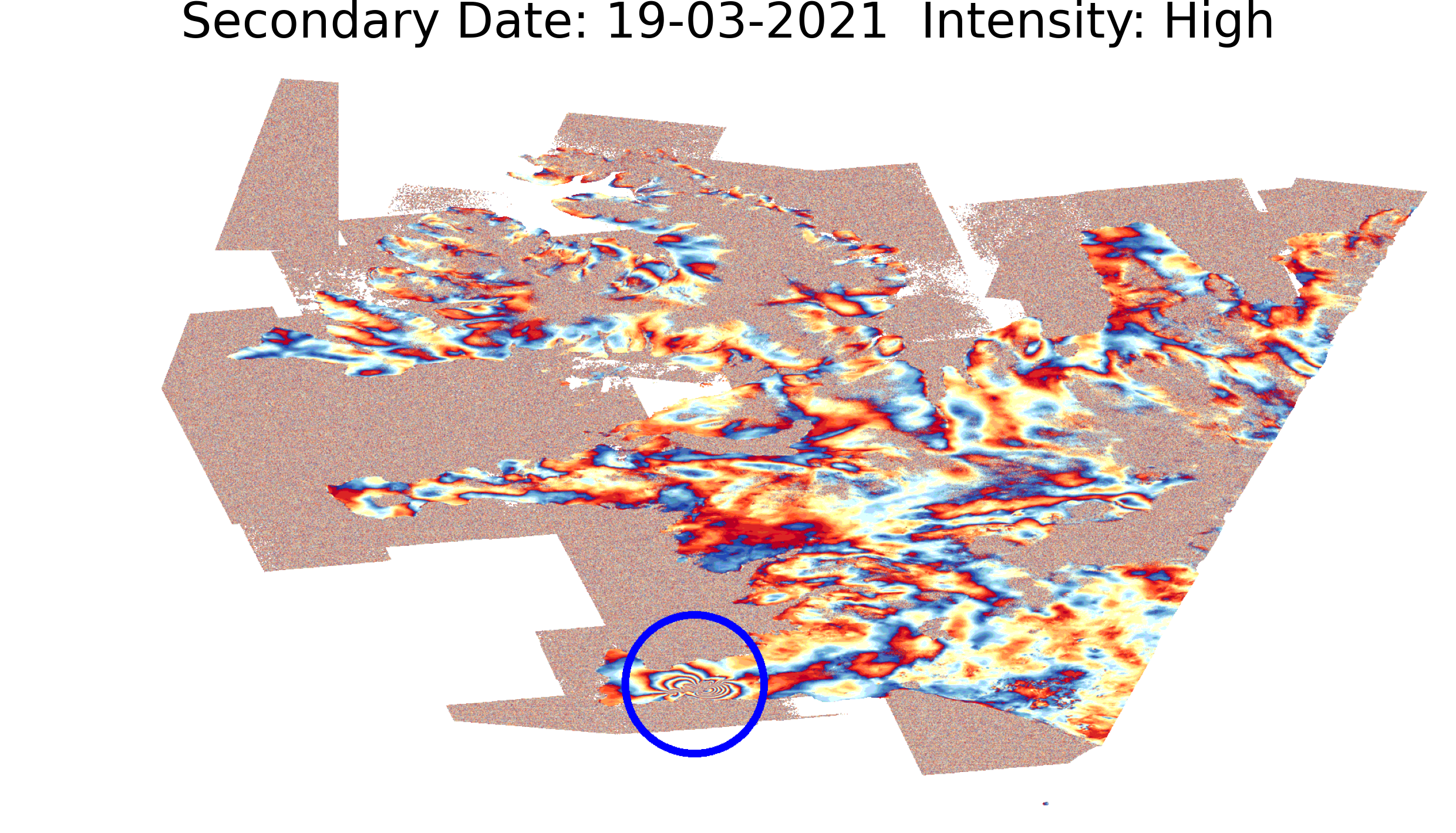}
    \end{subfigure}
        \begin{subfigure}{0.45\textwidth}
    \centering
    \includegraphics[width=\textwidth]{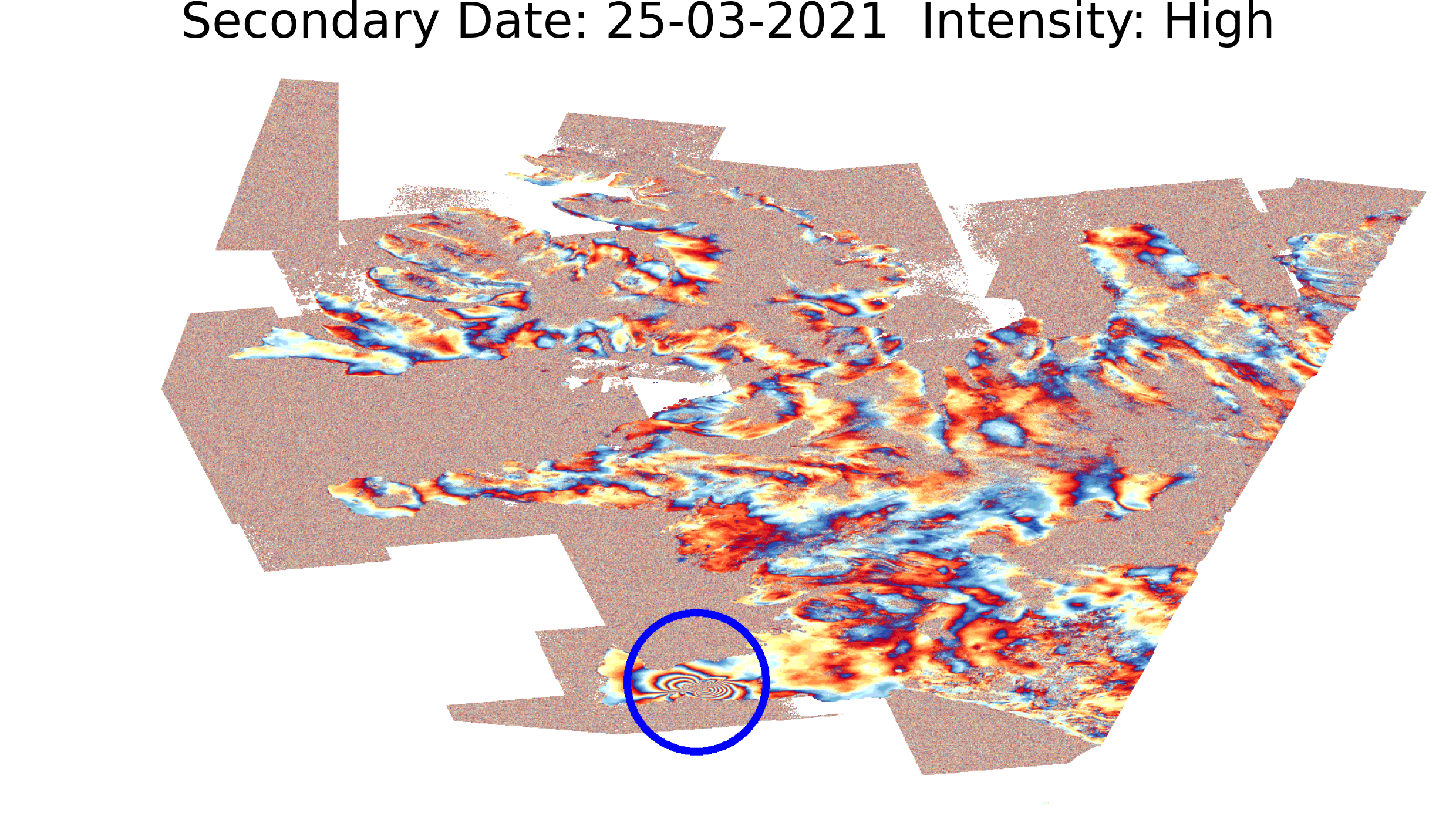}
    \end{subfigure}
    \caption{Fagradalsfjall volcano InSAR evolution with primary SAR date: 01-03-2021 and varying secondary date. The volcano erupted on 19-03-2021. The volcano and the respective deformation fringes can be found in the blue circle at the bottom left of the InSAR.}
    \label{fig:timeseries}
\end{figure*}

Each annotation file (~\cref{mint:json}) contains information regarding the spatial context of the InSAR frame (\textit{frameID}), as well as temporal information regarding the acquisition date of the primary and the secondary SAR images. Encoding the temporal component of the InSAR samples is a strong characteristic of \orionsar{}, which allows systematic volcano monitoring. Indicatively, in~\cref{fig:timeseries} we present the InSAR evolution of the recent eruption in Fagradalsfjall volcano, Iceland. In the time series, the primary date is fixed and the secondary date increments over time, while the eruption occurred at 19-03-2021~\cite{bountos2021self}. 
The InSAR time series of~\cref{fig:timeseries} show the expansion and magnification of the deformation fringes, as magma fills the volcanic chamber at depth with a certain rate. 

Naturally, the temporal information makes it easier to detect changes and therefore the onset of ground deformation. In fact the temporal information was exploited by the experts during the dataset labelling. However, the temporal ordering of the data to form a time-series is not straightforward, since the gap between the acquisition of the primary and the secondary SAR image is not fixed.

  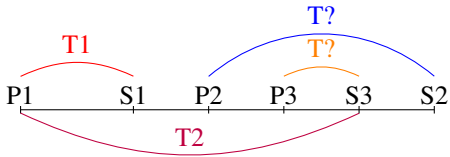
\begin{figure}[h]
      \centering
  \begin{tikzpicture}
    \draw (0 cm,0)  node [name=Primary1] {P1};
    \draw (1.5cm,0)  node [name=Secondary1] {S1};
    \draw (2.5 cm,0)  node [name=Primary2] {P2};
    \draw (5.5 cm,0) node [name=Secondary2] {S2};
    
    \draw ( 3.5cm, 0) node [name=Primary3] {P3};
    
    \draw (4.5cm,0) node [name=Secondary3] {S3};
    
    \draw (0,-0.2cm) -- (5.5,-0.2cm);
    
    \draw (0 cm,-0.25cm) -- (0cm, -0.15cm);
     \draw (1.5cm,-0.25cm) -- (1.5 cm, -0.15cm);
     \draw (2.5 cm,-0.25cm) -- (2.5 cm, -0.15cm);
  \draw (3.5 cm, -0.25cm) -- (3.5 cm, -0.15cm);
    
     \draw ( 5.5cm, -0.25cm) -- (5.5cm, -0.15cm);
    
    \draw (4.5cm,-0.25cm) -- (4.5cm,-0.15cm);

        \draw [red] (Primary1.north) to [bend left=25] node {} node [auto] {T1} (Secondary1.north);
    \draw [blue] (Primary2.north) to [bend left=40] node {} node [auto] {T?} (Secondary2.north);
    
    \draw [orange] (Primary3.north) to [bend left=25] node {} node [auto] {T?} (Secondary3.north);
    
    \draw [purple] (Primary1.south) to [bend right=25] node {} node [auto] {T2} (Secondary3.south);
  \end{tikzpicture}
  \caption{Illustration of the InSAR temporal ordering problem. P[1-3] stand for the primary images while S[1-3] for the secondary. The P1-S1 (T1) pair is clearly first in this sequence. The pair P1-S3 (T2) is second, as the
  the ordering is dictated by their secondary image. Finally, the pair P3-S3 pair is included in the P2-S2 pair, and there is no direct way to order them in this case.}
  \label{fig:ordering}
  \end{figure}

In this context we can identify three distinct cases for the creation of the time-series. Assuming two InSAR pairs A and B: first, if both the primary and the secondary dates of InSAR A are earlier than the respective dates of the InSAR B, then A goes before B and vice versa. Second, if both have the same primary date, the order is dictated by the secondary date of the two pairs. Third, if pair's B time-span is included in pair's A time-span, there is no intuitive way to order the pairs (see \cref{fig:ordering}). Randomly including one of the two in the time-series can result in loss of information due to varying InSAR quality. Therefore discovering an optimal, information-preserving, time series of interferograms is a machine learning task on its own.

\subsection{Preparation of a deep learning ready dataset}

\begin{figure}
\centering
\begin{subfigure}{0.07\textwidth}
   \includegraphics[width=\textwidth]{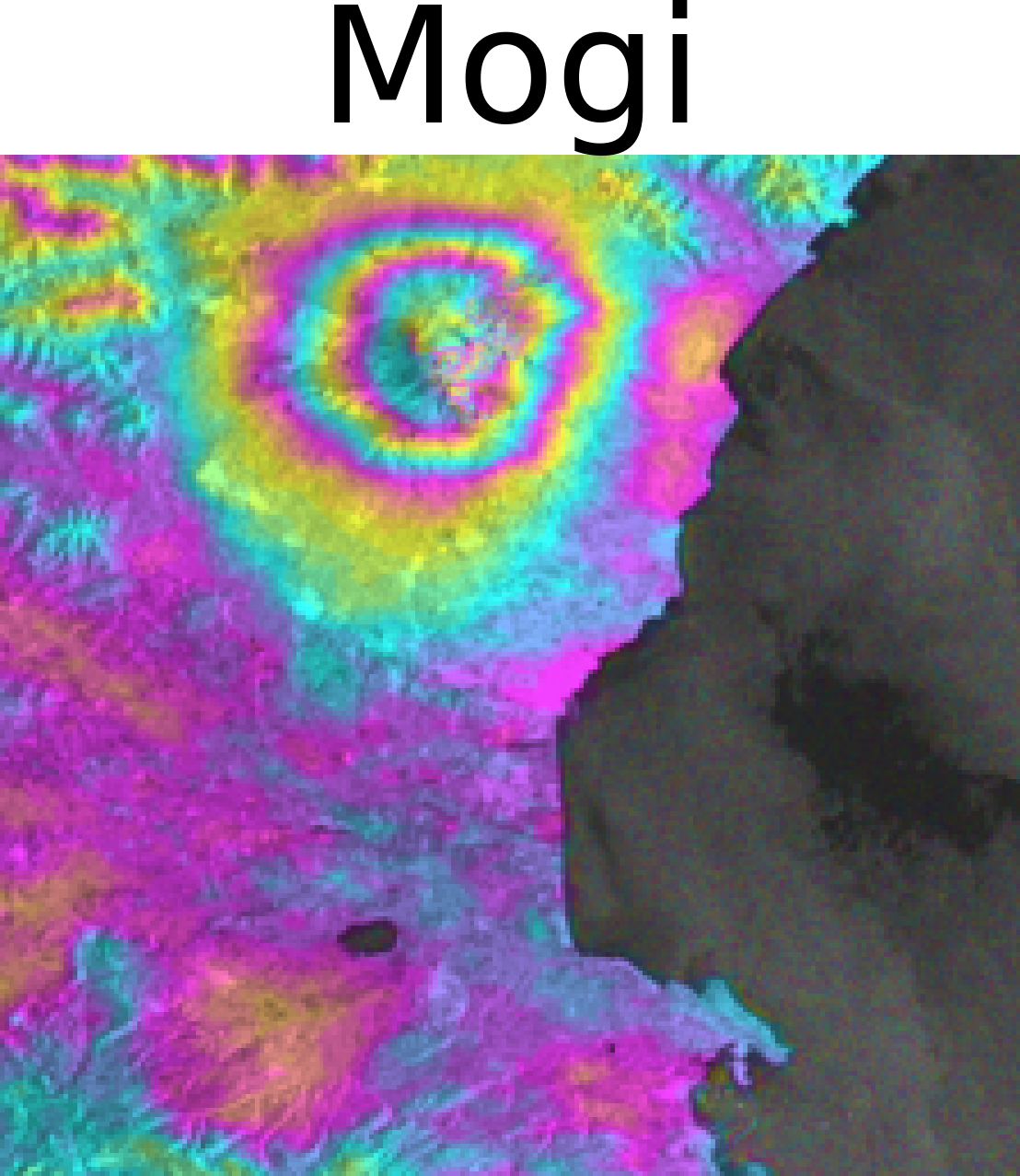} 
\end{subfigure}
\begin{subfigure}{0.07\textwidth}
   \includegraphics[width=\textwidth]{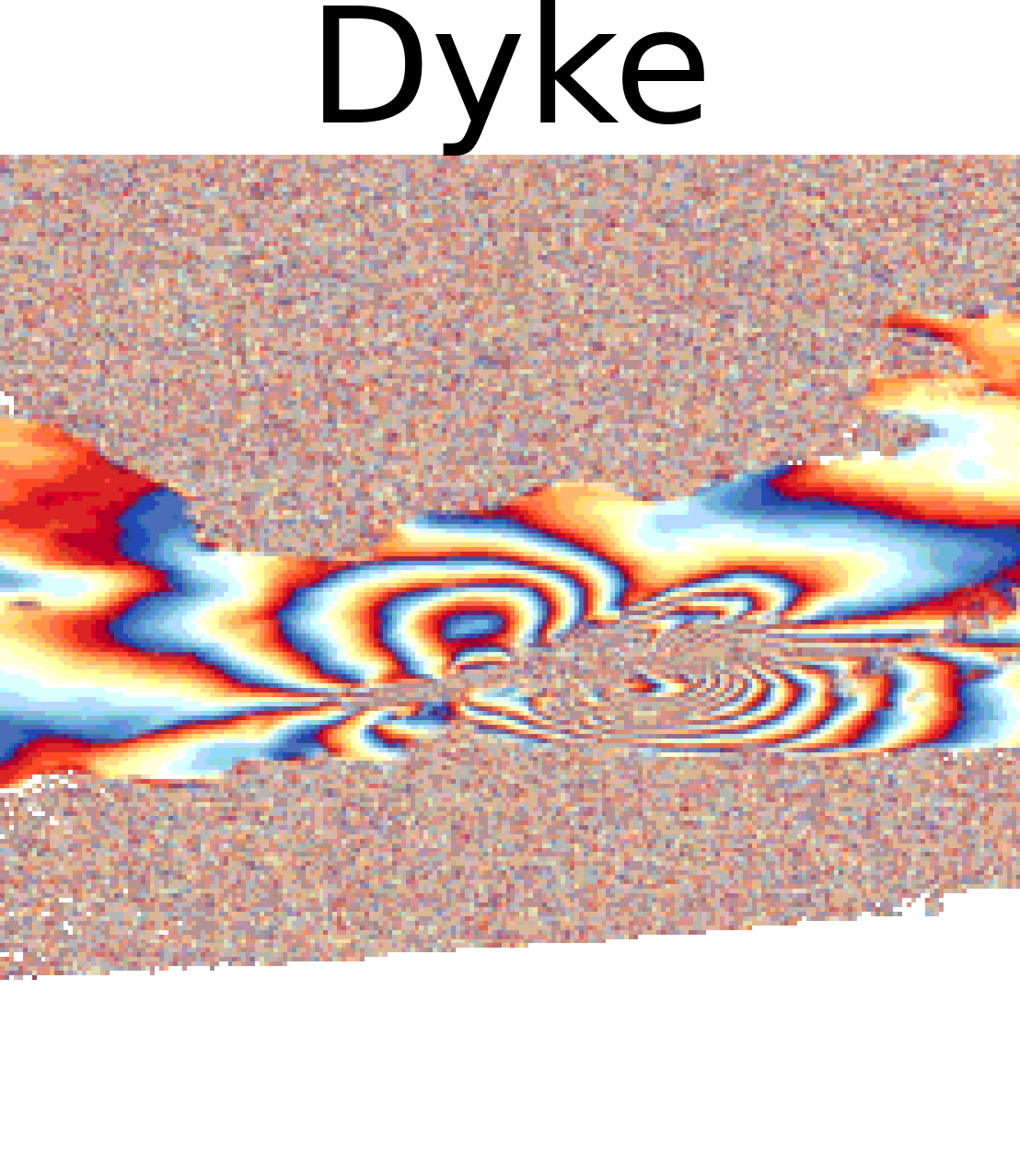} 
\end{subfigure}
\begin{subfigure}{0.07\textwidth}
   \includegraphics[width=\textwidth]{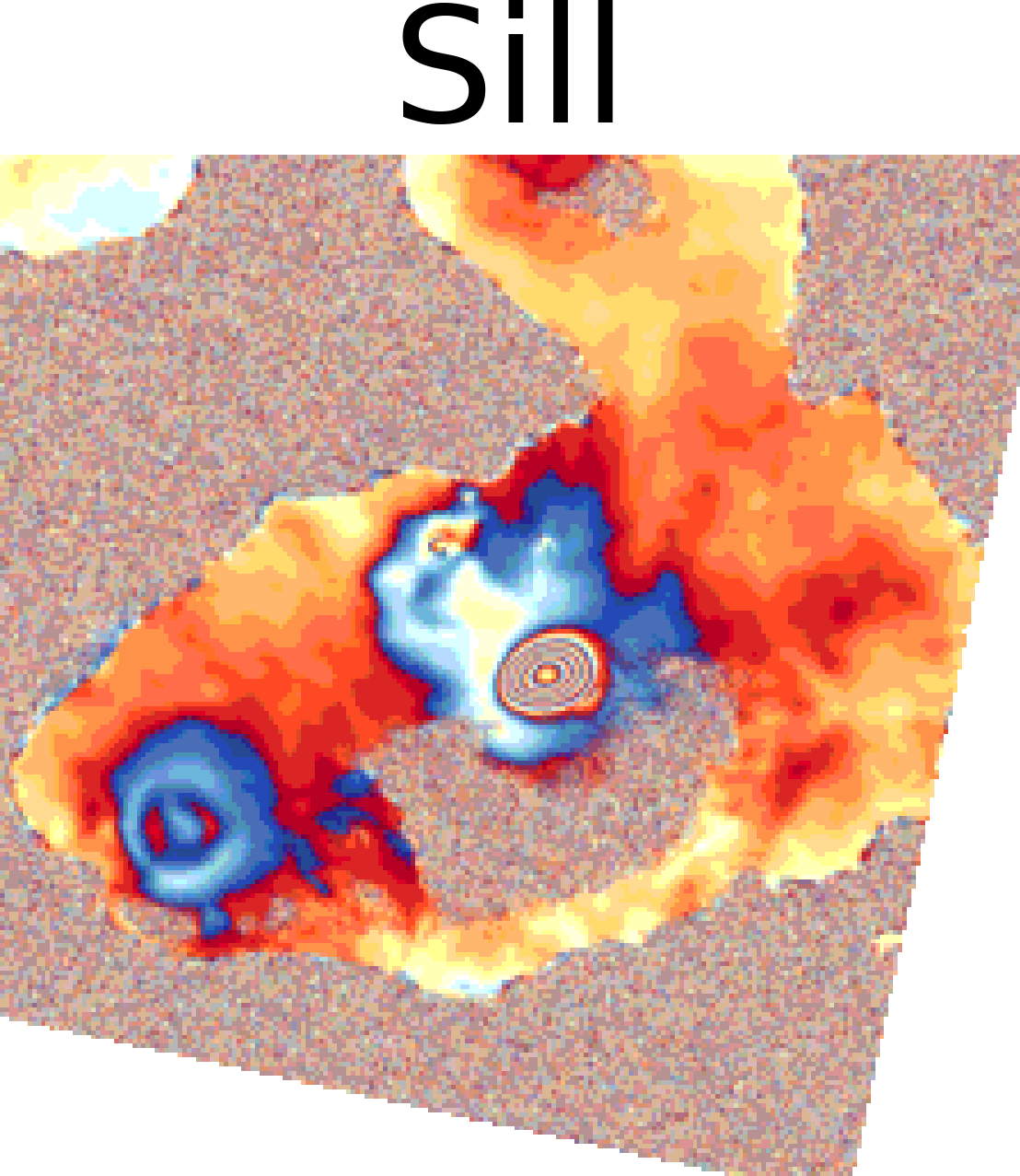} 
\end{subfigure}
\begin{subfigure}{0.07\textwidth}
   \includegraphics[width=\textwidth]{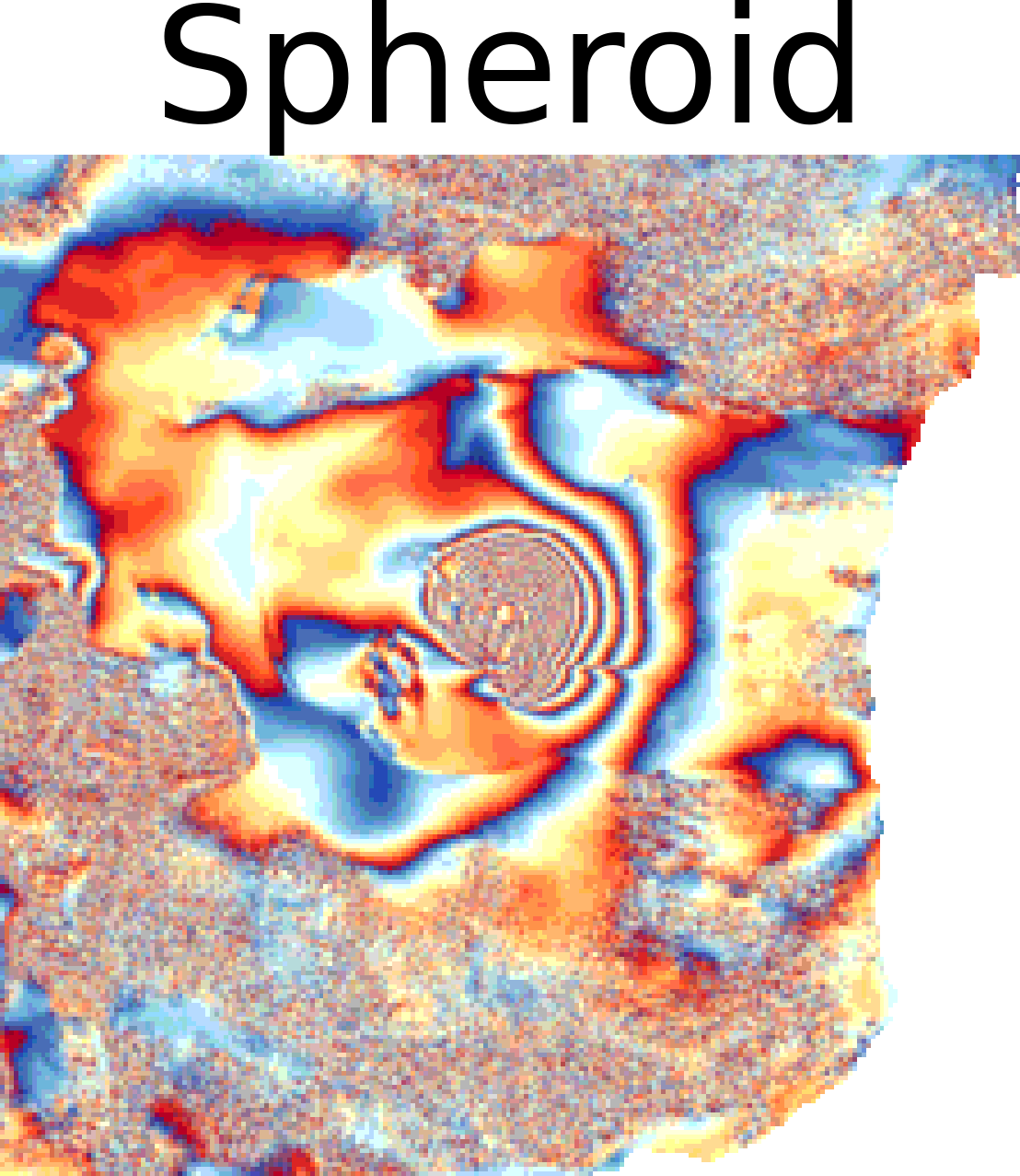} 
\end{subfigure}
\begin{subfigure}{0.07\textwidth}
   \includegraphics[width=\textwidth]{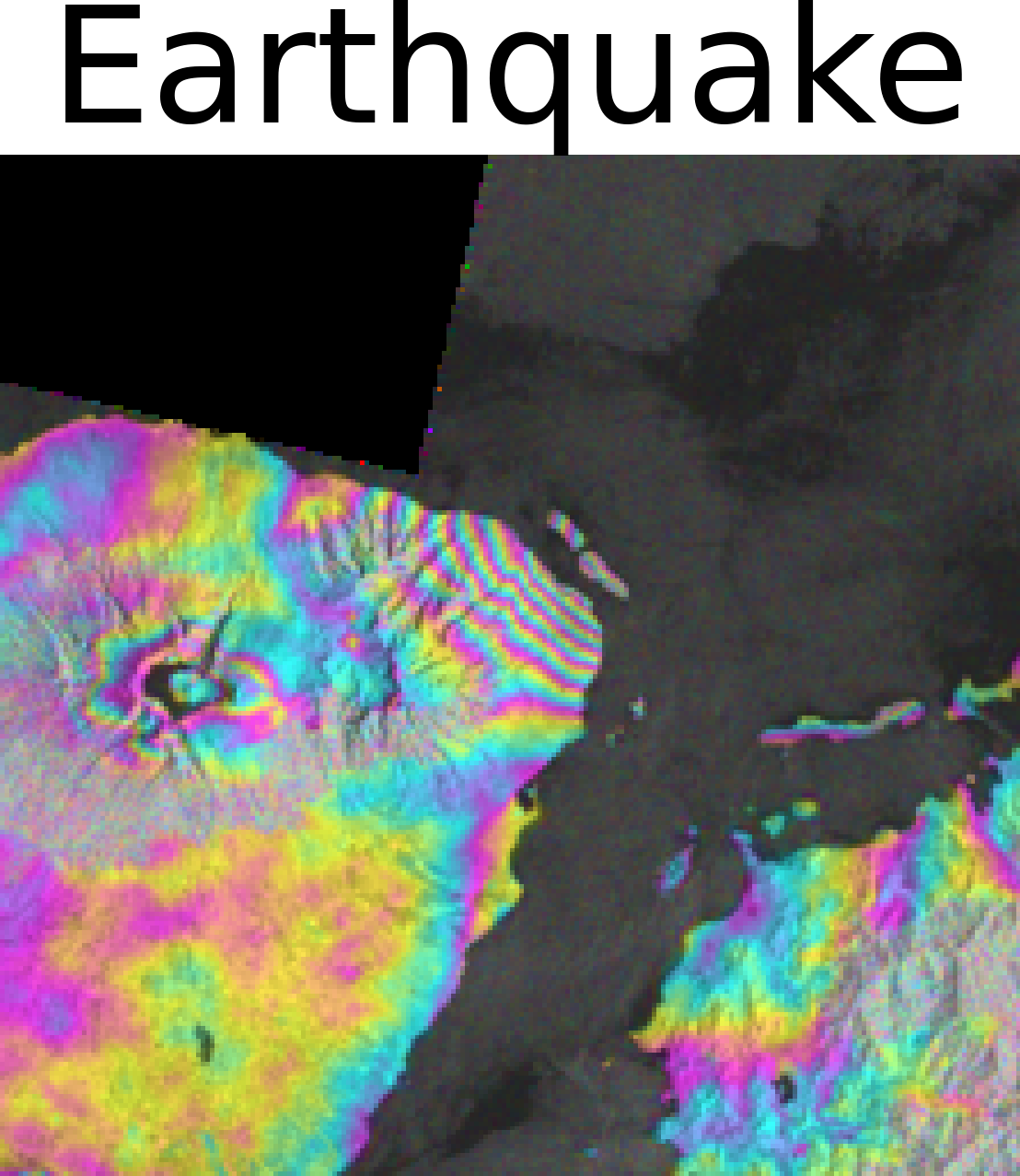} 
\end{subfigure}
\begin{subfigure}{0.07\textwidth}
   \includegraphics[width=\textwidth]{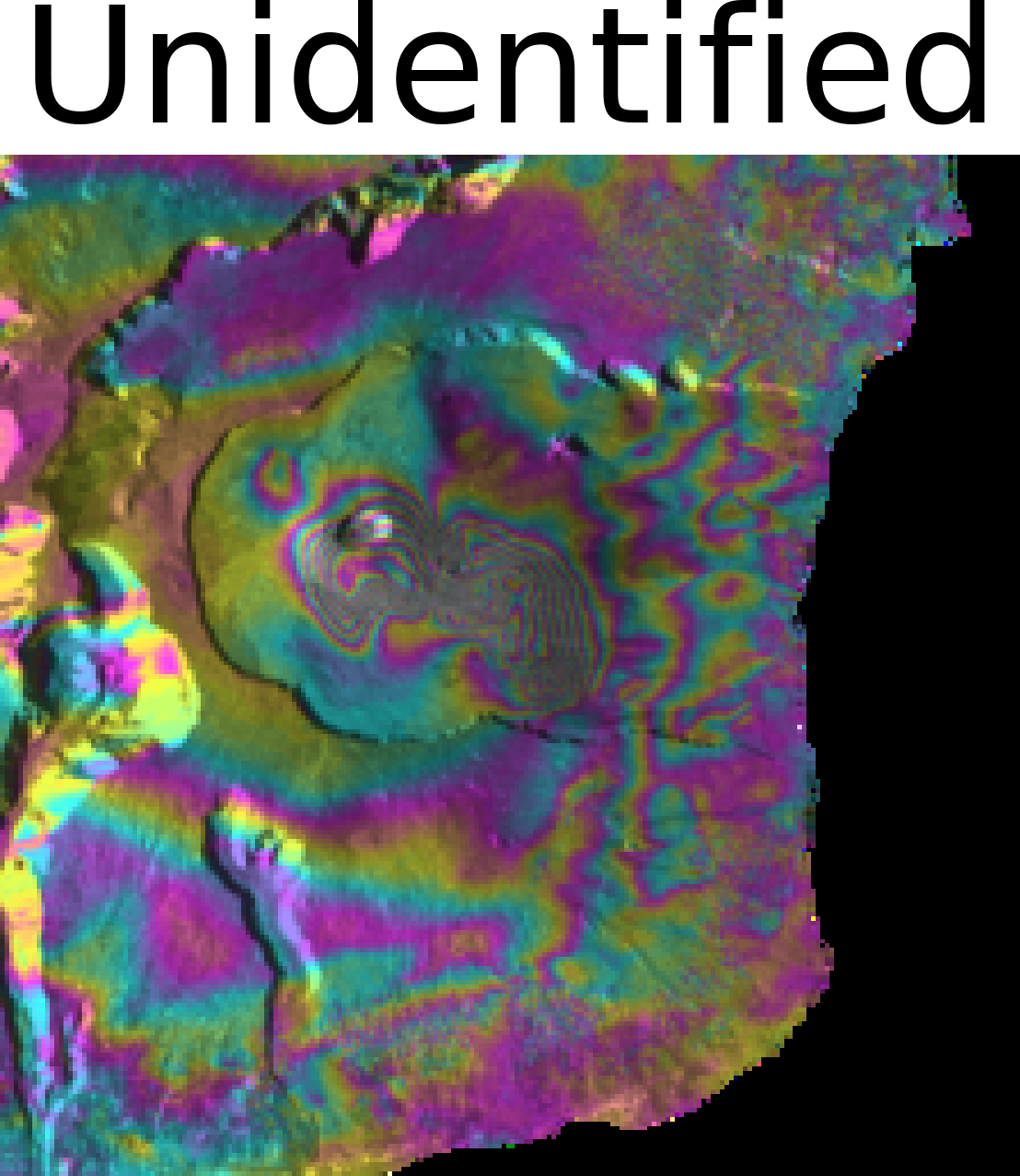} 
\end{subfigure}
\hfill
\begin{subfigure}{0.07\textwidth}
   \includegraphics[width=\textwidth]{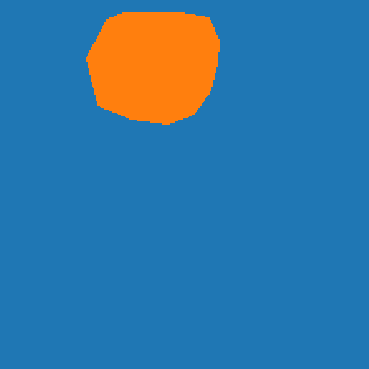} 
\end{subfigure}
\begin{subfigure}{0.07\textwidth}
   \includegraphics[width=\textwidth]{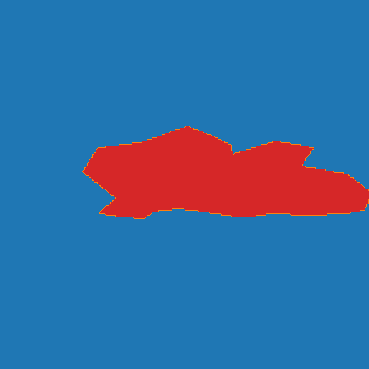} 
\end{subfigure}
\begin{subfigure}{0.07\textwidth}
   \includegraphics[width=\textwidth]{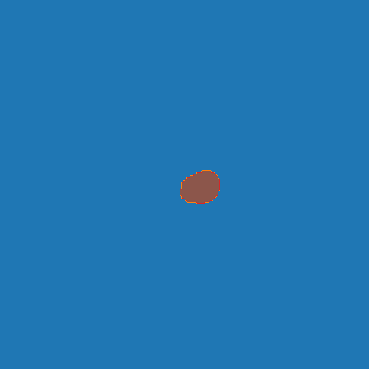} 
\end{subfigure}
\begin{subfigure}{0.07\textwidth}
   \includegraphics[width=\textwidth]{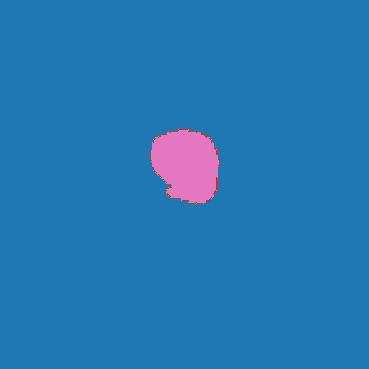} 
\end{subfigure}
\begin{subfigure}{0.07\textwidth}
   \includegraphics[width=\textwidth]{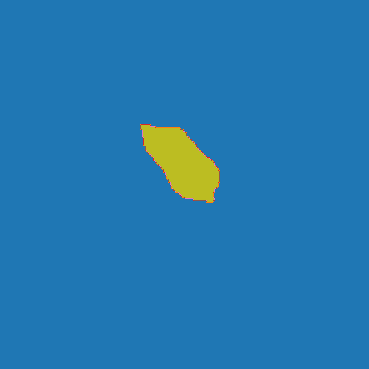} 
\end{subfigure}
\begin{subfigure}{0.07\textwidth}
   \includegraphics[width=\textwidth]{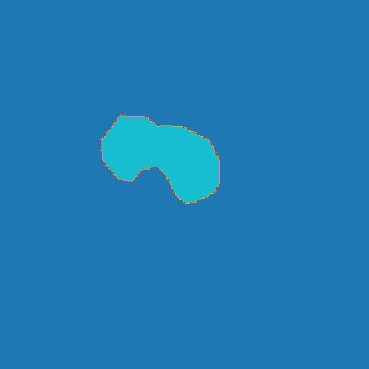} 
\end{subfigure}
          \caption{Examples of samples with different ground deformation types.}
          \label{fig:deformation_samples}
\end{figure}

\begin{figure}
     \centering
\begin{subfigure}{0.2\textwidth}
   \includegraphics[width=\textwidth]{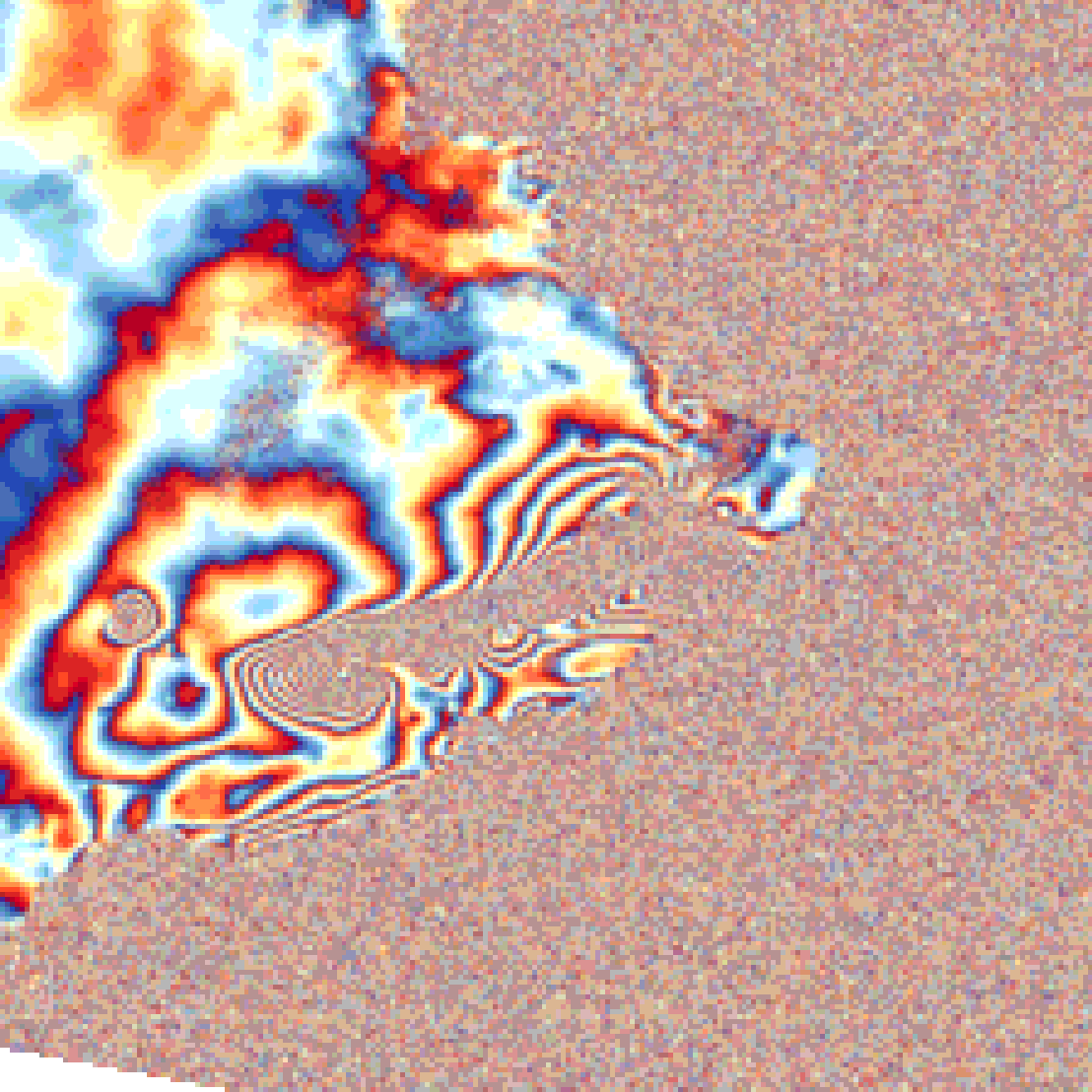}
\end{subfigure}
\begin{subfigure}{0.2\textwidth}
   \includegraphics[width=\textwidth]{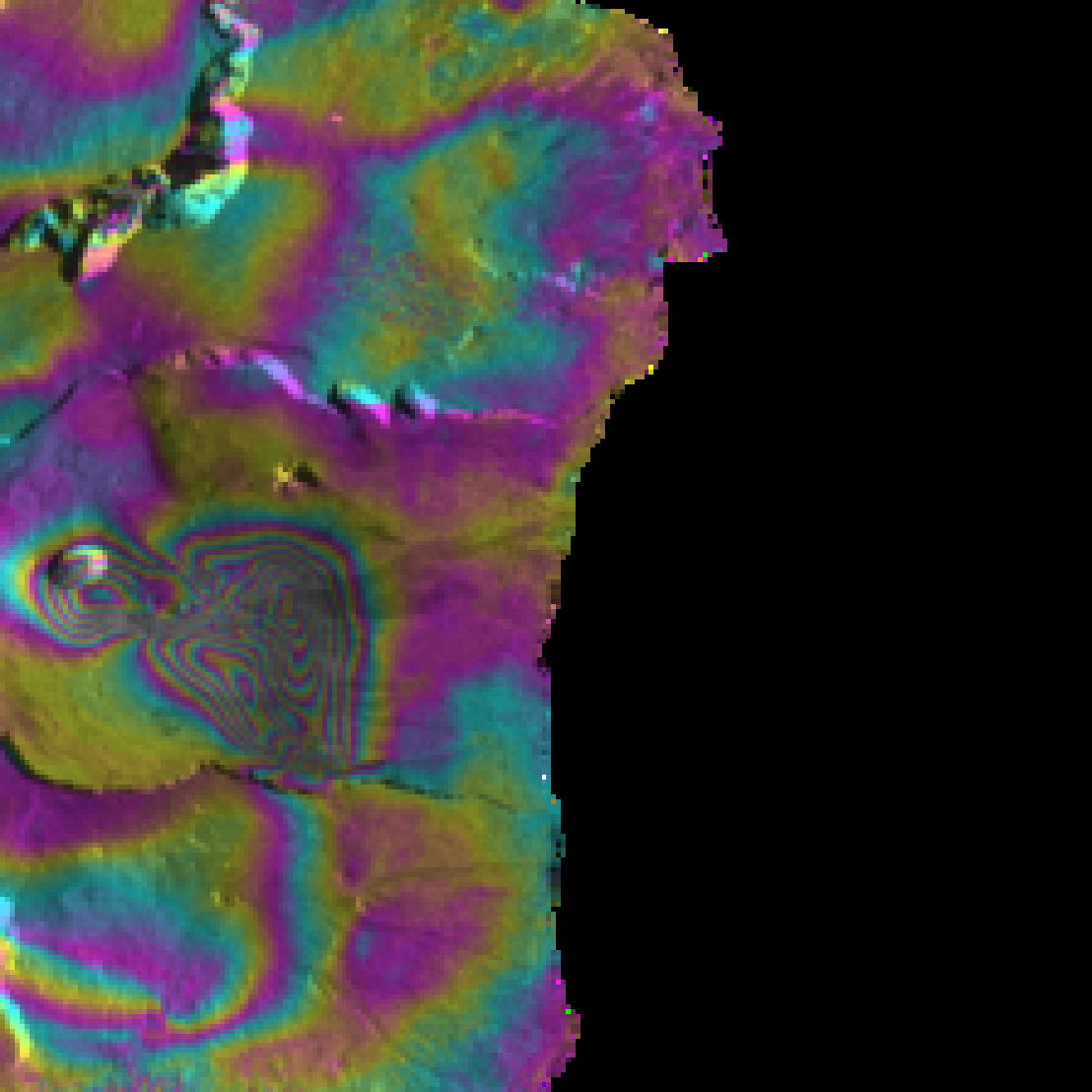}
\end{subfigure}
\hfill
\begin{subfigure}{0.2\textwidth}
   \includegraphics[width=\textwidth]{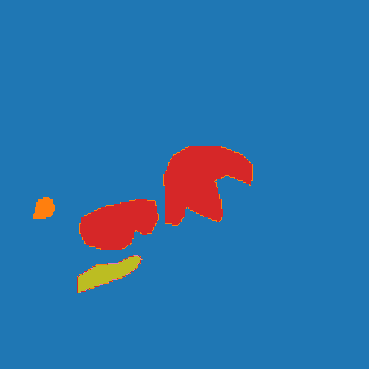}
\end{subfigure}
\begin{subfigure}{0.2\textwidth}
   \includegraphics[width=\textwidth]{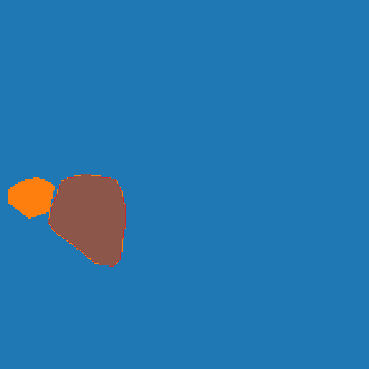}
\end{subfigure}
          \caption{Examples of samples containing multiple deformation fringes. The color code is the same as \cref{fig:deformation_samples}.}
          \label{fig:multilabel}
 \end{figure}

To create a deep learning ready dataset, we preprocess the frames and create patches with resolution of 224x224 pixels alongside the respective labels and segmentation masks. Each InSAR that contains ground deformation is randomly cropped to an area containing the deformation fringes. In the case of multiple deformation fringes in an interferogram, we create one patch for every instance without excluding their coexistence. The rest of the frames are broken down to a sequence of patches of the predefined resolution. The resulting dataset consists of 216,106 samples. 213,859 of them do not contain any deformation while 2,247 do. From the deformation samples there are 535 Mogi, 690 Dyke, 1,360 Sill, 28 Spheroid, 150 Earthquake and 50 Unidentified containing samples. Examples of these patches along with the respective masks can be seen in \cref{fig:deformation_samples}. Interestingly, some of the patches contain more than one type of volcanic deformation (\cref{fig:multilabel}). Therefore, this task can be expressed as a multi-class, multi-label DL problem.


\begin{table*}[ht]
\begin{tabular}{lllllllll}
      & No Deformation & Mogi & Dyke & Sill & Spheroid & Earthquake & Unidentified & Total Deformation Samples \\\toprule
Train & 198,253         & 211  & 294  & 522  & 20       & 35         & 50           & 1062                      \\\hline
Val   & 7803           & 163  & 204  & 409  & 6        & 60         & 0            & 592                       \\\hline
Test  & 7803           & 161  & 192  & 429  & 2        & 55         & 0            & 593\\\bottomrule
\end{tabular}
\caption{Distribution of the patches in the train, validation and test set.}\label{tab:valtest}
\end{table*}

\subsection{Dataset split} 
To setup a standard evaluation context, regarding ground deformation, we carefully split our dataset to train, validation and testing set. From a total of 38 frames we select 35 for training and three for validation and testing. Given the inherent class imbalance we split them in a way that preserves representatives of every deformation class in each set. 
The validation and test sets contain 826 deformation and 1648 non Deformation InSAR from the volcanoes of Sierra Negra, Cerro Azul, La Cumbre, Kilauea, Maune Loa, Puu Oo and Campi Flegrei. For tasks which operate on patches instead of full images, 
we randomly assign patches from these frames to the validation and test sets. The ground deformation distribution of the final split can be seen in \cref{tab:valtest}. We intentionally do not include the Unidentified class in the validation and test sets. This split should fit most tasks. In problems that require the entire InSAR \eg image captioning, we can use all InSAR with primary date until 2017 for validation and the rest for the test set. In specialized cases like glacier detection or total corruption detection, a more careful, task-oriented approach is needed.

\section{Computer vision tasks for InSAR understanding}
The multifaceted information of \orionsar{} dataset contains the necessary ingredients to model InSAR data. In this section we define a set of \orionsar{} inspired computer vision problems towards InSAR understanding.

\subsection{Volcanic deformation and activity classification}
\begin{figure}
    \centering
    \includegraphics[width=0.45\textwidth]{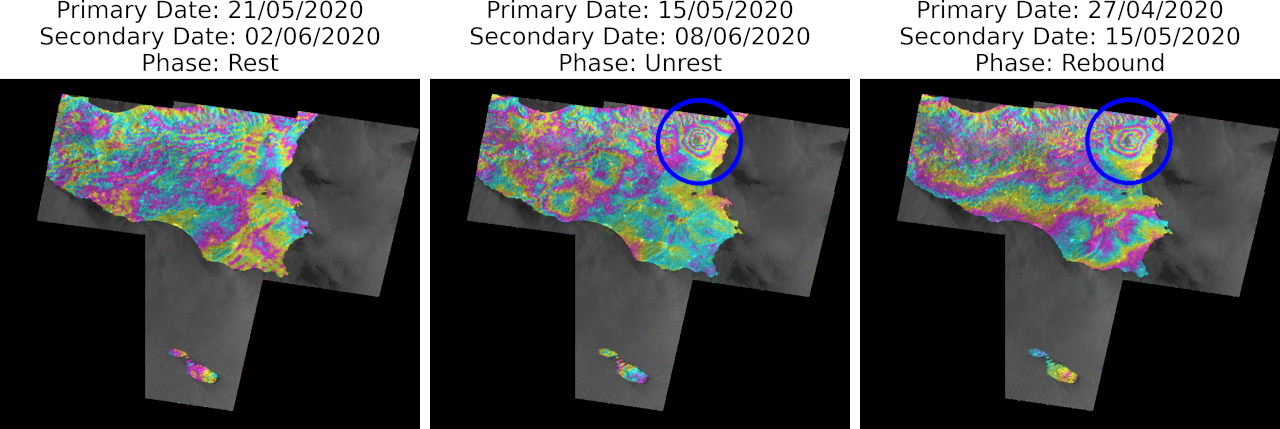}
    \caption{Samples from different phases of the Etna volcano.}
    \label{fig:phase_evolution}
\end{figure}

\begin{figure}
    \centering
    \includegraphics[width=0.45\textwidth]{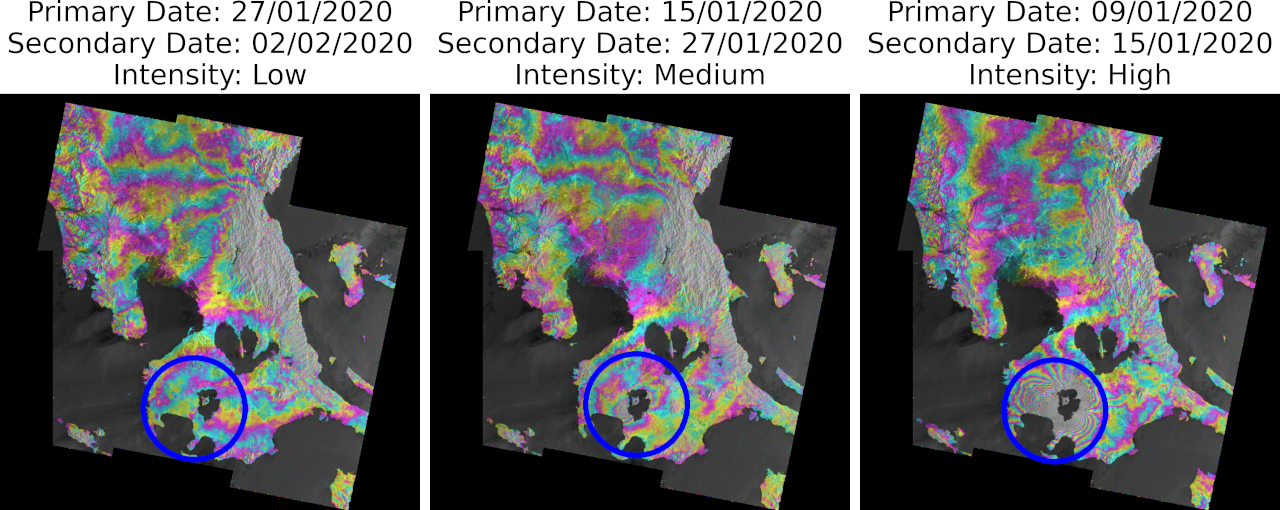}
    \caption{Samples with different intensity levels from the Taal volcano.}
    \label{fig:taal_intensity_evolution}
\end{figure}

\begin{figure}
    \centering
    \includegraphics[width=0.45\textwidth]{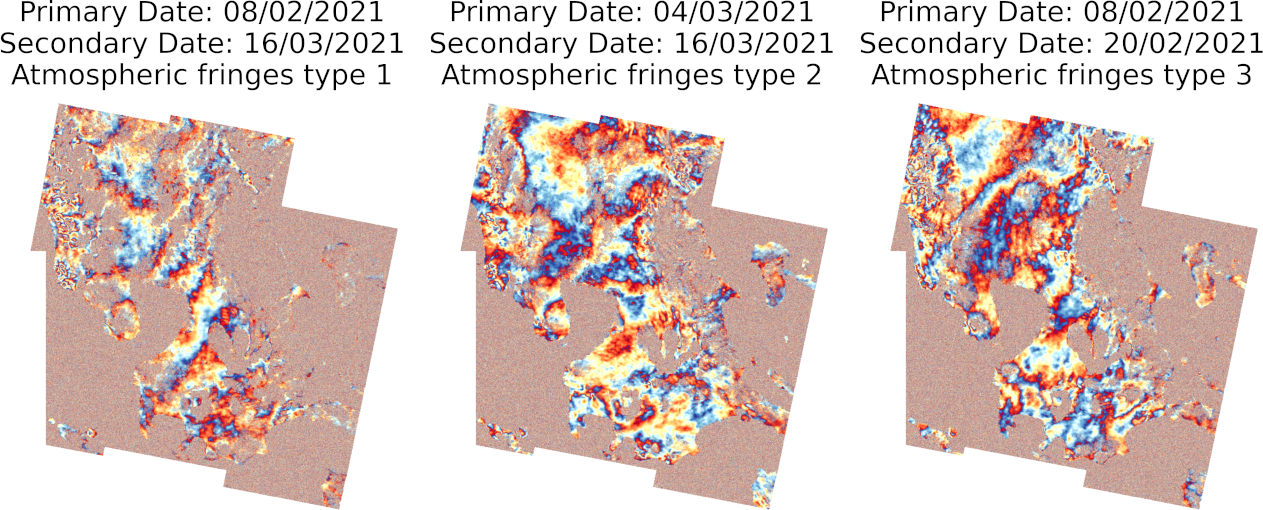}
    \caption{Samples with different types of atmospheric fringes.}
    \label{fig:taal_atmospheric_types}
\end{figure}

A fundamental task related to InSAR interpretation, under the umbrella of volcanic activity detection, is the classification of ground deformation and its specific patterns. 
\orionsar{} contains very rich information on ground deformation and atmospheric signals, with several categories that can co-exist.
Examples of the variations of event intensity, phase of the volcano and atmospheric fringes can be seen in \cref{fig:taal_intensity_evolution,fig:phase_evolution,fig:taal_atmospheric_types}. The problem can either be treated as binary, attempting to simply detect ground deformation or as a multi-class multi-label problem aiming to identify all major classes. However, for some of the classes in our study, \eg, the event intensity, the phase of the volcano and the type of the deformation, our problem transcends into a fine-grained image classification problem where we have to recognise subtle differences between similar classes. DL models can be assisted in this task, by using the segmentation mask that provides information on the location and extent of the deformation. Information regarding object location has already been used in \cite{yang2012unsupervised} and \cite{chai2013symbiotic}. A second option, not mutually exclusive, is to exploit the textual information~\cite{he2017fine} of the InSAR. In our case, however, the description refers to the entire InSAR and not to individual patches, which, for example, may not contain the general atmospheric contribution type that partially dominates the InSAR frame.  
A potential solution is to use an upscaled version of the entire InSAR as one sample or preprocess it with care to match the patch with the caption. 


\subsection{Semantic segmentation of deforming area}
The identification of the location and mask of the deformation can be of major help in volcano observatories, given the complexity of InSAR data, the small magnitude of the deformation signal in the early stages of volcanic unrest, as well as the difficulty to discriminate them in the presence of atmospheric and orbital errors clutter. To this end, we can utilize the provided segmentation masks for each of the deformation patches. Again, the problem can be treated as a binary or a multi-class problem attempting to classify the deformation and the specific deformation type respectively. 

\subsection{Text and InSAR}
\begin{figure}
    \centering
    \includegraphics[width=0.25\textwidth]{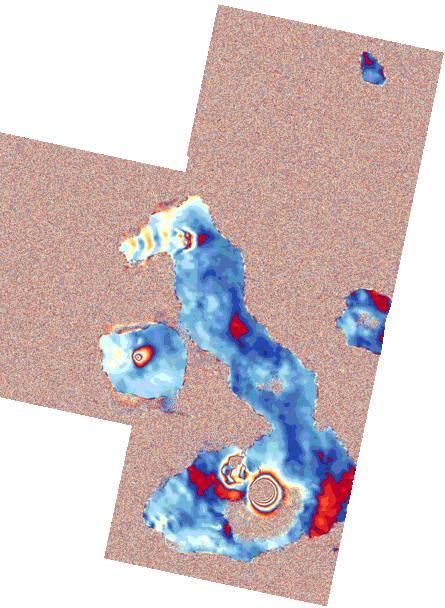}
    \caption{InSAR frame containing deformation. Caption: "Turbulent mixing effect or wave-like patterns caused by liquid and solid particles of the atmosphere can be detected around the area. High deformation activity can be detected."}
    \label{fig:annotation2}
\end{figure}

Text information has not been thoroughly exploited for remote sensing applications~\cite{tuia2021toward}. However, retrieving the semantic information of remote sensing imagery in textual form, especially for the complex InSAR data, can be of particular help for interpretation by non experts.
The availability of the text modality opens new directions towards remote sensing imagery understanding. Tasks related to such a crossmodal setting include image captioning, text to image generation for synthetic InSAR data generation and text based image retrieval. In our dataset we provide a detailed description of each InSAR frame we annotate (\cref{fig:annotation2}). 

\subsection{Quality assessment of InSAR imagery}
\begin{figure}
    \centering
    \includegraphics[width=0.45\textwidth]{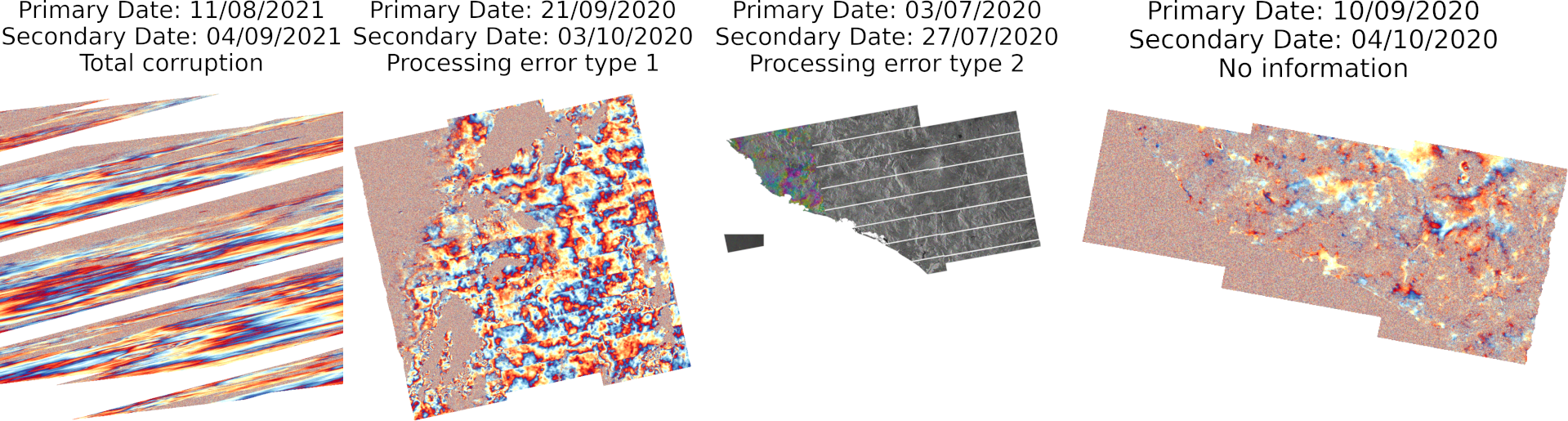}
    \caption{Representatives of different corruption flags.}
    \label{fig:corruption_types}
\end{figure}

In massive production systems like Comet-LiCS, it may happen that some faulty interferograms are generated. Detecting such InSAR as early as possible and reprocessing them is important. In \cref{fig:corruption_types} we selected some characteristic examples of such frames, which are labeled in \orionsar{} with a corruption related flag 
\eg \textit{no info}, \textit{low coherence}, \textit{processing error} and \textit{corrupted}, pointing directly to the existence of some sort of corruption. Similarly, the annotator's confidence contains critical meta information on the quality of the interferogram. Low confidence means the InSAR image does not provide clear information about its contents, making it difficult for an expert to interpret.

The quality related information contained in our annotation file can be exploited in two ways. First, by addressing the problem of faulty InSAR detection, which can be reduced to a classification scheme on the known flags i.e corrupted, processing errors, \etc. Second, by identifying information-poor interferograms. This task is directly connected to \cref{sec:spatiotemporal}, where we need to find the optimal (in regards to information) set of interferograms to describe the temporal evolution of ground deformation. This translates to selecting the most information-rich InSAR samples in order to construct the time series. We thus provide the coherence maps for each InSAR sample as a helping hand. We leave this as a task open to new solutions and metrics. 

\section{General InSAR representation learning}
\label{sec:ssl}

Self-supervised learning has gained a lot of traction in computer vision recently~\cite{chen2020simple,he2020momentum,Caron_2021_ICCV}. Given the abundance of unlabeled remote sensing data, and the range of crucial downstream tasks they can address, self-supervised learning is a particularly good fit. 
Ayush \etal~\cite{ayush2021geography} proposed a modification to the creation of positive pairs for contrastive learning exploiting the geographical context and the spatio-temporal structure of remote sensing imagery. Bountos \etal~\cite{bountos2021self} used self-supervised learning in the InSAR domain 
as a means to learn quality representations for volcanic unrest detection from highly imbalanced datasets. Class imbalance is a common encounter in nature and therefore in remote sensing, making it difficult to learn quality representations via supervised methods. \Cref{fig:class_distribution,tab:def_distr,tab:phase,fig:disturbances} highlight this for various categories in the \orionsar{} dataset. In a supervised learning context a deep learning model will make use of all kinds of underlying biases to correctly classify the patches refraining from learning general, transferable representations, applicable to other downstream tasks.
On the other hand, models trained in a self-supervised manner are more robust to class imbalance compared to supervised learning methods~\cite{liu2021self}. Evenmore, self-supervised learning in the wild can provide models with good few-shot capabilities~\cite{goyal2021self}. To this end, we have enhanced our dataset with random frames from all over the world, summing up to 384 unique frames and 110,573 InSAR in total, creating a large-scale InSAR dataset oriented towards self-supervised learning. Cropping to 224x224 resolution patches, the number of non-overlapping unlabeled training samples surpasses 1 million. 

Large datasets on their own are not necessarily enough to provide quality representations. Remote sensing data come with their own peculiarities, while current state of the art self-supervised learning methods tend to  rely heavily on hand-picked combinations of data augmentations tuned on optical data like ImageNet~\cite{deng2009imagenet}. It is yet unexplored, however, what kind of augmentations work best in the InSAR domain. With this dataset we can work towards the creation of general InSAR foundation models, while researching the application of self-supervised learning methods for InSAR data.

\section{Baseline experiments}
Motivated by the discussion in \cref{sec:ssl}, 
we train a ResNet18 using MoCo-v2\cite{he2020momentum} for 300 epochs. We followed the choices of Bountos \etal~\cite{bountos2021self} for the augmentation set (horizontal flip, vertical flip, cutout, elastic transformation, gaussian noise, gaussian blur and multiplicative noise)  and the initialization of the model with weights from ImageNet. This is the first large scale, InSAR model, trained in a self-supervised learning fashion, to be published, leaving the exploration for the optimal set of data augmentations for InSAR data as future work.

In addition, we address the binary version of the ground deformation classification problem, for which we provide three baseline models (\cref{tab:baselines}). 
For the first baseline we utilize a SwinPL model, pretrained on synthetic InSAR data~\cite{bountos2022learning}, as initialization and train it for 5 epochs.
We follow the original paper and use SGD with momentum. We set the learning rate to $0.001$, momentum to $0.9$ and weight decay to $0.0001$. 
For the second baseline, we use a ResNet-18 with the weights learned via MoCo-v2 for initialization. We denote this classifier as ResNet-18-MoCo. Finally, we compare these results with a simple ResNet-18 trained using the weights of ImageNet for initialization. Both ResNet-18 methods were trained for 5 epochs using the Adam optimizer with both learning rate and weight decay set to $0.0001$. We use oversampling in all our experiments to create balanced batches.

\begin{table}[]
\begin{tabular}{lllll}
Model          & ACC & P & R & F \\\toprule
SwinPL         &97.3\% & 76.9\%& 88.5\% & 82.2\% \\\hline
ResNet-18      & 97.4\%  & 79.9\%  & 84.6\%  &  82.2\% \\\hline
ResNet-18-MoCo & 97.3\% & 89.1\% & 71.8\% & 79.5\% \\\bottomrule
\end{tabular}
\caption{Results for ground deformation classification. ACC stands for Accuracy, P for Precision, R for Recall and F for F-Score.}
\label{tab:baselines}
\end{table}
\section{Conclusion}
In this work, we present \orionsar{}, the first large scale Sentinel-1 InSAR dataset that was manually annotated by a team of experts. The dataset is inspired by global volcanic unrest detection, but was designed to address multiple computer vision tasks related to InSAR interpretation, including image classification, semantic segmentation, and image captioning. Finally, we provide some baseline deep learning models for ground deformation classification. We believe that \orionsar{} will pave the way to further exploit the rich archive of InSAR imagery and spur new machine learning applications for geohazards mitigation.   

\section*{Acknowledgment}
This work has received funding from the European Union’s Horizon 2020 research and innovation project DeepCube, under grant agreement number 101004188. LiCSAR contains modified Copernicus Sentinel data [2014-2021] analysed by the Centre for the Observation and Modelling of Earthquakes, Volcanoes and Tectonics (COMET). LiCSAR uses JASMIN, the UK’s collaborative data analysis environment (http://jasmin.ac.uk).
\clearpage

{\small
\bibliographystyle{ieee_fullname}
\bibliography{bibliography}
}

\end{document}